\documentclass[final]{cvpr}

\usepackage{times}
\usepackage{epsfig}
\usepackage{graphicx}
\usepackage{amsmath}
\usepackage{amssymb}
\usepackage{multirow}
\usepackage{multicol}
\usepackage{booktabs}
\usepackage{appendix}

\usepackage[pagebackref=False,breaklinks=true,colorlinks,bookmarks=false]{hyperref}

\def\cvprPaperID{10827} 
\def\confYear{CVPR 2021}

\begin{document}

\title{3D Object Detection with Pointformer}

\author{%
  Xuran Pan$^{1}$\thanks{Equal contribution.}\ \ \
  Zhuofan Xia$^{1}$\footnotemark[1]\ \ \
  Shiji Song$^{1}$\ \ \
  Li Erran Li$^{2}$\thanks{Work done prior to Amazon.}\ \ \
  Gao Huang$^{1}$\thanks{Corresponding author.}\\
    $^{1}$Department of Automation, Tsinghua University, Beijing, China\\
    Beijing National Research Center for Information Science and Technology (BNRist),\\
    $^{2}$Alexa AI, Amazon / Columbia University\\
  \texttt{\small \{pxr18, xzf20\}@mails.tsinghua.edu.cn}, \texttt{\small{} erranlli@gmail.com},\\
  \texttt{\small\{shijis, gaohuang\}@tsinghua.edu.cn}
}

\maketitle

\begin{abstract}
Feature learning for 3D object detection from point clouds is very challenging due to the irregularity of 3D point cloud data. In this paper, we propose Pointformer, a Transformer backbone designed for 3D point clouds to learn features effectively. Specifically, a Local Transformer module is employed to model interactions among points in a local region, which learns context-dependent region features at an object level. A Global Transformer is designed to learn context-aware representations at the scene level. To further capture the dependencies among multi-scale representations, we propose Local-Global Transformer to integrate local features with global features from higher resolution. In addition, we introduce an efficient coordinate refinement module to shift down-sampled points closer to object centroids, which improves object proposal generation. We use Pointformer as the backbone for state-of-the-art object detection models and demonstrate significant improvements over original models on both indoor and outdoor datasets. Code and pre-trained models are available at \href{https://github.com/Vladimir2506/Pointformer}{https://github.com/Vladimir2506/Pointformer}.

\end{abstract}
\section{Introduction}
3D object detection in point clouds is essential for many real-world applications such as autonomous driving \cite{Geiger2012KITTI} and augmented reality \cite{Park2008Multiple3O}. Compared to images, 3D point clouds can provide detailed geometry and capture 3D structure of the scene. On the other hand, point clouds are irregular, which can not be processed by powerful deep learning models, such as convolutional neural networks directly. This poses a big challenge for effective feature learning.

The common feature processing methods in 3D detection can be roughly categorized into three types, based on the form of point cloud representations. \textit{Voxel-based} approaches \cite{Song2016CVPR, Hou2019CVPR, CBGS2019} gridify the irregular point clouds into regular voxels and are followed by sparse 3D convolutions to learn high dimensional features. Though effective, voxel-based approaches face the dilemma between efficiency and accuracy. Specifically, using smaller voxels gains more precision, but suffers from higher computational cost. Conversely, using larger voxels misses potential local details in the crowded voxels.

\begin{figure}
    \begin{center}
    \begin{tabular}{cc}
        Image of the scene &
        Ground truth  \\
        \includegraphics[width=0.25\linewidth]{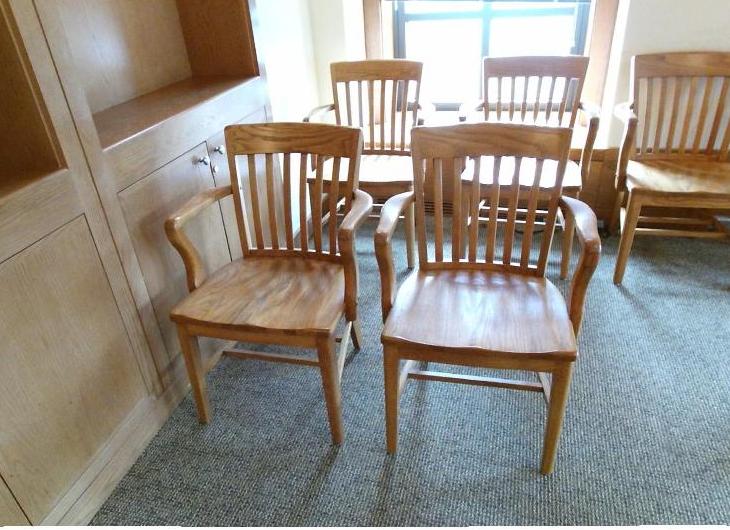} &
        \includegraphics[width=0.3\linewidth]{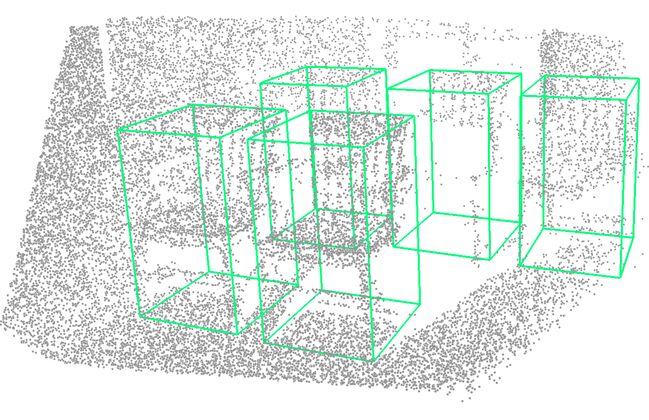} \\
    \end{tabular}
    \begin{tabular}{ccc}
        Top-50 attention &
        Top-100 attention &
        Top-200 attention \\
        \includegraphics[width=0.28\linewidth]{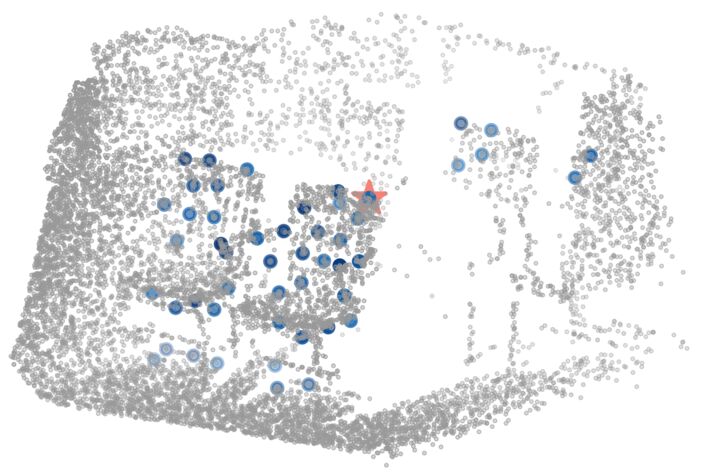} &
        \includegraphics[width=0.28\linewidth]{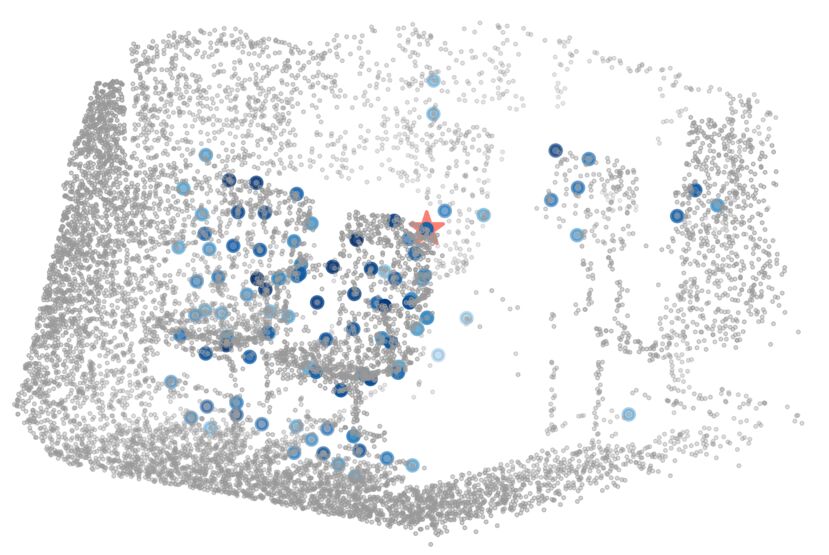} &
        \includegraphics[width=0.28\linewidth]{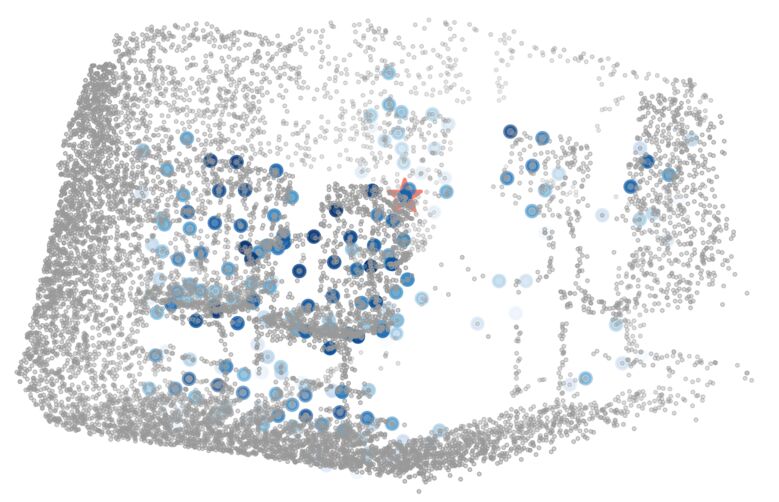} \\
    \end{tabular}
    \caption{\textbf{Attention maps directly from \textit{Pointformer} block, darker blue indicates stronger attention.}  
    For the key point (star), \textit{Pointformer} first focuses on the local region of the same object (the back of the chair), then spreads the attention to other regions (the legs), finally attends to points from other objects globally (other chairs), leveraging both local and global dependencies.}
    \end{center}
    \label{fig:1}
\end{figure}

\begin{figure*}
    \begin{center}
    \includegraphics[width=0.9\linewidth]{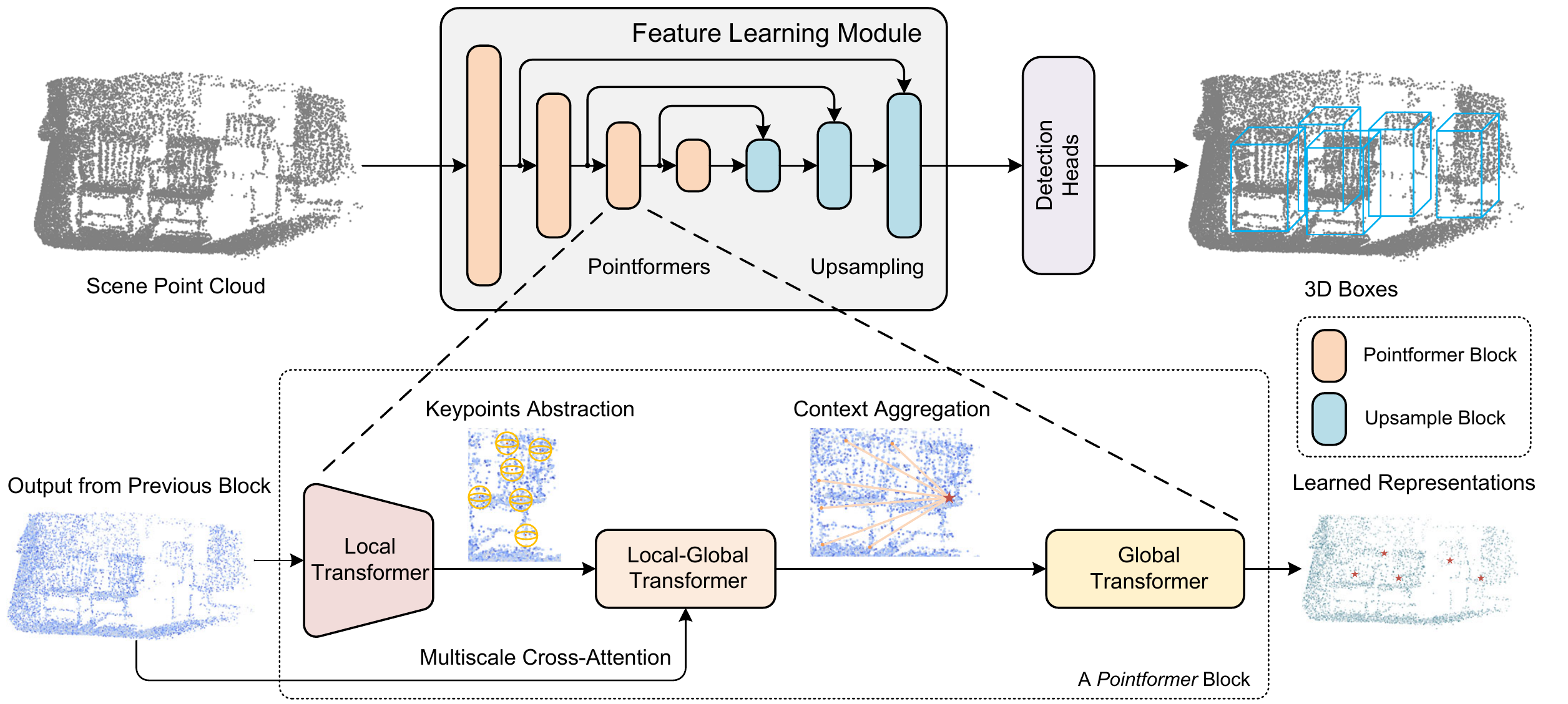}
    \caption{The \textbf{Pointformer} backbone for 3D object detection in point clouds. A basic feature learning block consists of three parts: a Local Transformer to model interactions in the local region; a Local-Global Transformer to integrate local features with global information; a Global Transformer to capture context-aware representations at the scene level.}
    \label{fig:2}
    \end{center}
\end{figure*}

Alternatively, \textit{point-based} approaches \cite{Shi2019PointRCNN}, inspired by the success of PointNet \cite{qi2016pointnet} and its variants, consume raw points directly to learn 3D representations, which mitigates the drawback of converting point clouds to some regular structures. Leveraging learning techniques for point sets, point-based approaches avoid voxelization-induced information loss and take advantage of the sparsity in point clouds by only computing on valid data points. Nevertheless, due to the irregularity of point cloud data, point-based learning operations have to be permutation-invariant and adaptive to the input size.
To achieve this, it learns simple symmetric functions (e.g. using point-wise feedforward networks with pooling functions) which highly restricts its representation power.

\textit{Hybrid approaches} \cite{Zhou2018VoxelNet, lang2019Pointpillars, Ye2020HVNetHV, Shi2020PVRCNNPF} attempt to combine both voxel-based and point-based representations. \cite{Zhou2018VoxelNet, lang2019Pointpillars} leverages PointNet features at the voxel level and a column of voxels (pillar) level respectively. \cite{Ye2020HVNetHV, Shi2020PVRCNNPF} deeply integrate voxel features and PointNet features at the scene level. However, the fundamental difference between the two representations could pose a limit on the effectiveness of these approaches for 3D point-cloud feature learning.

To address the above limitations, we resort to the Transformer \cite{vaswani2017attention} models, which have achieved great success in the field of natural language processing. Transformer models \cite{devlin-etal-2019-bert} are very effective at learning context-dependent representations and capturing long range dependencies in the input sequence. Transformer and the associate self-attention mechanism not only meet the demand of permutation invariance, but also are proved to be highly expressive. Specifically, \cite{cordonnier2019relationship} proves that self-attention is at least as expressive as convolution. Currently, self-attention has been successfully applied to classification \cite{Ramachandran2019StandAloneSI} and 2D object detection \cite{Carion2020detr} in computer vision. However, the straightforward application of Transformer to 3D point clouds is prohibitively expensive because computation cost grows quadratically with the input size.

To this end, we propose \textit{Pointformer}, a backbone for 3D point clouds to learn features more effectively by leveraging the superiority of the Transformer models on set-structured data. 
As shown in Figure~\ref{fig:2}, Pointformer is a U-Net structure with multi-scale Pointformer blocks. A Pointformer block consists of Transformer-based modules that are both expressive and friendly to the 3D object detection task. First, a Local Transformer (LT) module is employed to model interactions among points in the local region, which learns context-dependent region features at an object level. Second, a coordinate refinement module is proposed to adjust centroids sampled from Furthest Point Sampling (FPS) which improves the quality of generated object proposals. Third, we propose Local-Global Transformer (LGT) to integrate local features with global features from higher resolution. Finally, Global Transformer (GT) module is designed to learn context-aware representations at the scene level. As illustrated in Figure \ref{fig:1}, Pointformer can capture both local and global dependencies, thus boosting the performance of feature learning for scenes with multiple cluttered objects.

Extensive experiments have been conducted on several detection benchmarks to verify the effectiveness of our approach. We use the proposed Pointformer as the backbone for three object detection models, CBGS \cite{CBGS2019}, VoteNet \cite{qi2019deep}, and PointRCNN \cite{Shi2019PointRCNN}, and conduct experiments on three indoor and outdoor datasets, SUN-RGBD \cite{song2015sun}, KITTI \cite{Geiger2012KITTI}, and nuScenes \cite{Caesar2020nuScenesAM} respectively. We observe significant improvements over the original models on all experiment settings, which demonstrates the effectiveness of our method. 

In summary, we make the following contributions:
\begin{itemize}
\item We propose a pure transformer model, \textit{Pointformer}, which serves as a highly effective feature learning backbone for 3D point clouds. Pointformer is permutation invariant, local and global context-aware. 

\begin{figure*}
    \begin{center}
    \includegraphics[width=0.9\linewidth]{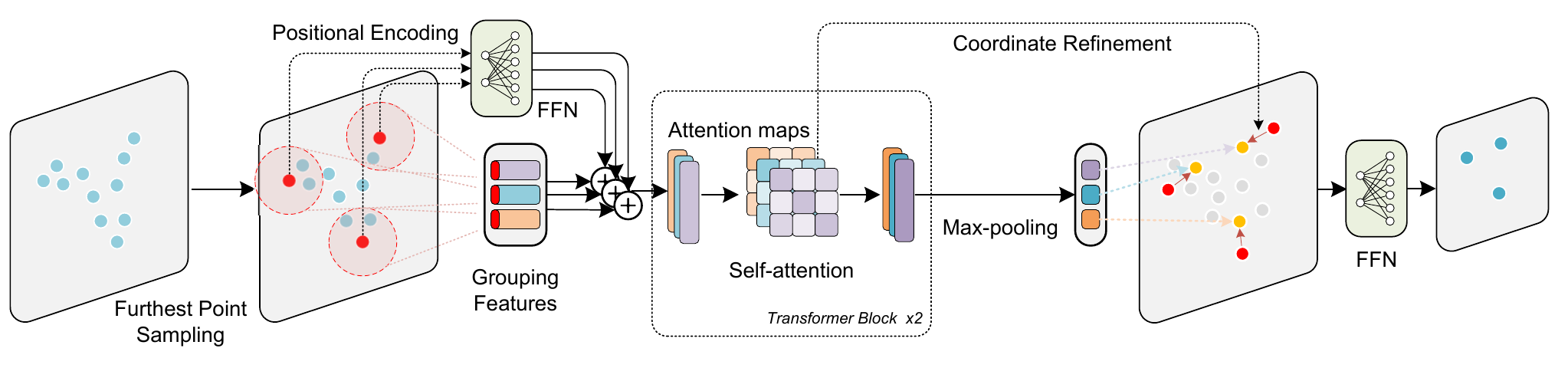}
    \caption{Illustration of the Local Transformer. Input points are first down-sampled by FPS and generate local regions by ball query. Transformer block takes point features and coordinates as input and generate aggregated features for the local region. To further adjust the centroid points, attention maps from the last Transformer layer are adopted for coordinate refinement. As a result, points are pushed closer to the object centers instead of surfaces.}
    \label{fig:3}
    \end{center}
\end{figure*}

\item We show that Pointformer can be easily applied as the drop-in replacement backbone for state-of-the-art 3D object detectors for the point cloud. 
\item We perform extensive experiments using Pointformer as the backbone for three state-of-the-art 3D object detectors, and show significant performance gains on several benchmarks including both indoor and outdoor datasets. This demonstrates that the versatility of Pointformer as 3D object detectors are typically designed and optimized for either indoor or outdoor only.
\end{itemize}
\section{Related Work}
\noindent
\textbf{Feature learning for 3D point clouds.} 
Prior work includes feature learning on voxelized grids, direct feature learning on point clouds and the hybrid of the two. 3D sparse convolution \cite{Graham2018sparse3dconv} is very effective on voxel grids. For direct feature learning, PointNet \cite{qi2016pointnet} and PointNet++ \cite{qi2017pointnet++} learn point-wise features and region features using feed-forward networks and simple symmetric functions (e.g. max) respectively. PCCN \cite{Wang2018contconv} generalizes convolution to non-grid structured data by exploiting parameterized kernel functions that span the full continuous vector space. EdgeConv \cite{wang2019DGCNN} exchanges local neighborhood information and acts on graphs dynamically computed in each layer of the network. Hybrid methods combine both types of features at the local level \cite{Zhou2018VoxelNet, lang2019Pointpillars}  or at the network level \cite{Ye2020HVNetHV, Shi2020PVRCNNPF}.

\noindent
\textbf{Transformers in computer vision.} Image GPT \cite{chen2020imageGPT} is the first to adopt the Transformers in 2D image classification task for unsupervised pretraining. Further, ViT \cite{dosovitskiy2020ViT} extends this scheme to large scale supervised learning on images. For high level vision tasks, DETR \cite{Carion2020detr} and Deformable DETR \cite{zhu2020deformable} leverage the advantages of Transformers in 2D object detection. Set Transformer \cite{lee2019setTransformer} uses attention mechanisms to model interactions among elements in the input set. In the field of 3D vision, PAT \cite{PAT2019CVPR} designs novel group shuffle attentions to capture long range dependencies in point clouds. To the best of our knowledge, we are the first to propose a pure Transformer model for 3D points clouds feature learning with carefully designed Transformer blocks and a positional encoding module to capture geometric and rich context information.

\noindent
\textbf{3D object detection in point clouds.} Detectors are designed either with point clouds as the only input \cite{CBGS2019, Zhou2018VoxelNet, lang2019Pointpillars, Ye2020HVNetHV, Shi2020PVRCNNPF, Shi2019PointRCNN, qi2019deep, shi2020point, xie2020mlcvnet, zhang2020h3dnet} or fusing multiple sensor modalities such as LiDAR and camera \cite{qi2018frustum, Liang2019msensor, Vora2020pointpainting}. Their backbones are designed with the aforementioned feature learning approaches. We focus on point cloud only object detection. In this category, VoxelNet \cite{Zhou2018VoxelNet} divides the point cloud into voxels, followed by 3D convolutions to extract features. VoteNet \cite{qi2019deep} devises a novel 3D proposal mechanism using deep Hough voting, before H3DNet \cite{zhang2020h3dnet} makes further investigations on geometric primitives. In addition, MLCVNet \cite{xie2020mlcvnet} focuses more on contextual information aggregation based on VoteNet, and PointGNN \cite{shi2020point} exploits graph learning methods in point cloud detection. 
We show that our novel Transformer based model, Pointformer, can be used as a drop-in replacement for voxel-based detector, CBGS \cite{CBGS2019} and point-based detectors, VoteNet \cite{qi2019deep} and PointRCNN \cite{Shi2019PointRCNN}.

\section{Pointformer}
Feature learning for 3D point clouds needs to confront its irregular and unordered nature as well as its varying size. Prior work utilizes simple symmetric functions, e.g., point-wise feedforward networks with pooling functions \cite{qi2016pointnet,qi2017pointnet++}, or resorts to the techniques in graph neural networks by aggregating information from the local neighborhood \cite{wang2019DGCNN}. However, the former is not effective in incorporating local context-dependent features beyond the capability of the simple symmetric functions; the latter focuses on the message passing between the center point and its neighbors while neglecting the feature correlations among the neighbor points. Additionally, global representations are also informative but rarely used in 3D object detection tasks.

In this paper, we design Transformer-based modules for point set operations which not only increase the expressiveness of extracting local features, but incorporate global information into point representations as well. As shown in Figure \ref{fig:2}, a Pointformer block mainly consists of three parts: Local Transformer (LT), Local-Global Transformer (LGT) and Global Transformer (GT). For each block, LT first receives the output from its previous block (high resolution) and extracts features for a new set with fewer elements (low resolution). Then, LGT uses the multi-scale cross-attention mechanism to integrate features from both resolutions. Lastly, GT is adopted to capture context-aware representations. As for the up-sampling block, we follow PointNet++ and adopt the feature propagation module for its simplicity.

\subsection{Background}
We first revisit the general formulation of the Transformer model. Let $F\!=\!\{f_i\}$ and $X\!=\!\{x_i\}$ denote a set of input features and their positions, where $f_i$ and $x_i$ represent the feature and position of token $i$, respectively. Then, a Transformer block comprises of a multi-head self-attention module and feedforward network:
\begin{equation}
    \label{b1}
    q_i^{(m)} = f_iW_q^{(m)}, k_i^{(m)} = f_iW_k^{(m)}, v_i^{(m)} = f_iW_v^{(m)},
\end{equation}
\begin{equation}
    y_i^{(m)} = \sum_{j}\sigma(q_i^{(m)}k_j^{(m)}/\sqrt{d} + {\rm PE}(x_i, x_j))v_j^{(m)},
\end{equation}
\begin{equation}
    y_i = f_i + {\rm Concat}(y_i^{(0)},y_i^{(1)},\dots,y_i^{(M-1)}),
\end{equation}
\begin{equation}
    \label{b4}
    o_i = y_i + {\rm FFN}(y_i),
\end{equation}
where $W_q, W_k, W_v$ are projections for query, key and value. $m$ is the index of $M$ attention heads and $d$ is the feature dimension. ${\rm PE(\cdot)}$ is the positional encoding function for input positions, and ${\rm FFN(\cdot)}$ represents a position-wise feed-forward network. $\sigma(\cdot)$ is a normalization function and \textit{SoftMax} is mostly adopted.

In the following sections, for simplicity, we use
\begin{equation}
    O = {\rm Transblock}(F, {\rm PE}(X)),
\end{equation}
to represent the basic Transformer block (Eq.(\ref{b1})$\sim$ Eq.(\ref{b4})). Readers can refer to \cite{vaswani2017attention} for further details.

\subsection{Local Transformer}
\label{local}

In order to build a hierarchical representation for a point cloud scene, we follow the high level methodology to build feature learning blocks on different resolutions \cite{qi2017pointnet++}. Given an input point cloud $\mathcal{P}={\{x_1,x_2,\dots,x_N\}}$, we first use furthest point sampling (FPS) to choose a subset of points $\{x_{c_1},x_{c_2},\dots,x_{c_{N'}}\}$ as a set of centroids. For each centroid, ball query is applied to generate $K$ points in the local region within a given radius. Then we group these features around the centroids, and feed them as a point sequence to a Transformer layer, as shown in Figure \ref{fig:3}. Let $\{x_i, f_i\}_t$ denote the local region for $t_{\rm th}$ centroid, where $x_i \in \mathbb{R}^{3}$ and $f_i \in \mathbb{R}^{C}$ represent the coordinates and features of the $i$-th points in the group, respectively. Subsequently, a shared $L$-layer Transformer block is applied to all local regions which receives the input of $\{x_i, f_i\}_t$ as follows:
\begin{align}
	\label{pe1}
   &f_i^{(0)} ={\rm FFN}(f_i), \ \forall i\in\mathcal{N}(x_{c_t}), \\
    F^{(l+1)}\! =\! &{\rm Transblock}(F^{(l)}, {\rm PE}(X)), l\!=\!0,..,L-1,
\end{align}
where $F=\{f_i|i\in\mathcal{N}(x_{c_t})\}$ and $X=\{x_i|i\in\mathcal{N}(x_{c_t})\}$ denote the set of features and coordinates in the local region with centroid $x_{c_t}$. 

Compared to the existing local feature extraction modules in \cite{Xu2020GridGCNFF, Yan2020PointASNLRP, Thomas2019KPConvFA}, the proposed Local Transformer has several advantages. First, the dense self-attention operation in the Transformer block greatly enhances its expressiveness. Several graph learning based approaches can be approximated as special cases of the LT module with learned parameter space carefully designed. For instance, a generalized graph feature learning function can be formulated as:
\begin{align}
    e_{ij} &= {\rm FFN}({\rm FFN}(x_i \oplus x_j) + {\rm FFN}(f_i\oplus f_j)),\\
    &f_i'= \mathcal{A}(\sigma(e_{ij})\times{\rm FFN}(f_j), \forall j \in \mathcal{N}(x_i)),
\end{align}
where most of the models utilize summation as the aggregation function $\mathcal{A}$ and the operation $\oplus$ is chosen from \{Concatenation, Plus, Inner-product\}. Therefore, the edge function $e_{ij}$ is at most a quadratic function of $\{x_i, x_j, f_i, f_j\}$. For a one-layer Transformer block, the learning module can be formulated with the inner-product self-attention mechanism as follows:
\begin{align}
    \label{query1}
    e_{ij}& = \frac{f_iW_{q}W_{k}^Tf_j^T}{\sqrt d} + {\rm PE}(x_i, x_j),\\
    f_i'= \mathcal{A}&(\sigma(e_{ij}\times{\rm FFN}(f_j), \forall j \in \mathcal{N}(x_i)),
\end{align}
where $d$ is the feature dimension of $f_i$ and $f_j$. We can observe that the edge function is also a quadratic function of $\{x_i, x_j, f_i, f_j\}$. With sufficient number of layers in ${\rm FFN}$s, the graph-based feature learning module has the same expressive power as a one-layer Transformer encoder. When it comes to \textit{Pointformer}, as we stack more Transformer layers in the block, the expressiveness of our module is further increased and can extract better representations.

Moreover, feature correlations among the neighbor points are also considered, which are commonly omitted in other models. Under some circumstances, neighbor points can be even more informative than the centroid point. Therefore, by leveraging message passing among all points, features in the local region are equally considered, which makes the local feature extraction module more effective. 

\subsection{Coordinate Refinement}
Furthest point sampling (FPS) is widely used in many point cloud frameworks, as it can generate a relatively uniform sampled points while keeping the original shape, which ensures that a large fraction of the points can be covered with limited centroids. However, there are two main issues in FPS: (1) It is notoriously sensitive to the outlier points, leading to highly instability especially when dealing with real-world point clouds. (2) Sampled points from FPS are a subset of original point clouds, which makes it challenging to infer the original geometric information in the cases that objects are partially occluded or not enough points of an object are captured.
Considering that points are mostly captured on the surface of objects, the second issue may become more critical as the proposals are generated from sampled points, resulting in a natural gap between the proposal and ground truth.

To overcome the aforementioned drawbacks, we propose a point coordinate refinement module with the help of the self-attention maps. As shown in Figure \ref{fig:3}, we first take out the self-attention map of the last layer of the Transformer block for each attention head. Then, we compute the average of the attention maps and utilize the particular row for the centroid point as a weight vector:
\begin{equation}
    W=\frac{1}{M}\sum_{m=1}^MA^{(m)}_{0,:},
\end{equation}
where $M$ represents the number of attention heads and $A^{(m)}$ is the attention map for the $m_{\rm th}$ head. Lastly, the refined centroid coordinates are computed as weighted average of all points in the local region:
\begin{equation}
    x_{c_t}' = \sum_{k=1}^{K}w_kx_k,
\end{equation}
where $w_k$ is the $k_{\rm th}$ entry of $W$. With the proposed coordinate refinement module, centroid points are adaptively moving closer to object centers. Moreover, by utilizing the self-attention map, our module introduces little computational cost and no additional learning parameters, making the refinement process more efficient.

\subsection{Global Transformer}
Global information representing scene contexts and feature correlations between different objects is also valuable in the detection tasks. Prior work using PointNet++ \cite{qi2017pointnet++} or sparse 3D convolution to extract high level features for 3D point clouds enlarges the receptive field as the depth of their networks increases. However, this has limitations on modeling long-range interactions. 

As a remedy, we leverage the power of Transformer modules on modeling non-local relations and propose a Global Transformer to achieve message passing through the whole point cloud. Specifically, all points are gathered to a single group $\mathcal{P}$ and serves as input to a Transformer module. The formulation for GT is summarized as follows:
\begin{equation}
    \label{pegt}
    f_i^{(0)} ={\rm FFN}(f_i), \ \forall i \in \mathcal{P},
\end{equation}
\begin{equation}
    F^{(l+1)}\! =\! {\rm Transblock}(F^{(l)},\! {\rm PE}(X)), l\!=\!0,..,L-1.
\end{equation}

By leveraging the Transformer on the scene level, we can capture the context-aware representations and promote message passing among different objects. Moreover, global representations can be particularly helpful for detecting objects with very few points.

\begin{equation}
    {\rm PE}(x_i, x_j)={\rm FFN}(x_i-x_j).
\end{equation}

\subsection{Local-Global Transformer}
Local-Global Transformer is also a key module to combine the local and global features extracted by the LT and GT modules. As shown in Figure \ref{fig:2}, the LGT adopts a multi-scale cross-attention module and generates relations between low resolution centroids and high resolution points. Formally, we apply cross attention similar to the encoder-decoder attention used in Transformer. The output of LT serves as query and the output of GT from the higher resolution is used as key and value. With the $L$-layer Transformer block, the module is formulated as:
\begin{equation}
    \label{pe3}
    f^{(0)} = {\rm FFN}(f_i), \forall i \in \mathcal{P}^l,
\end{equation}
\begin{equation}
    f_j' = {\rm FFN}(f_j), \forall j \in \mathcal{P}^h,
\end{equation}
\begin{equation}
    F^{(l\!+\!1)}\! =\! {\rm Transblock}(F^{(l)}\!, F_j', {\rm PE}(X)),\! l\!=\!0,\!..,\!L\!-\!1,
\end{equation}
where $\mathcal{P}^l$ (keypoints, the output of LT in Figure \ref{fig:2}) and $\mathcal{P}^h$ (the input of a Pointformer block in Figure \ref{fig:2}) represent subsamples of point cloud $\mathcal{P}$ from low and high resolution respectively.
Through the Local-Global Transformer module, we utilize whole centroid points to integrate global information via an attention mechanism, which makes the feature learning of both more effective.

\subsection{Positional Encoding}
Positional encoding is an integral part of Transformer models as it is the only mechanism that encodes position information for each token in the input sequence. When adapting Transformers for 3D point cloud data, positional encoding plays a more critical role as the coordinates of point clouds are valuable features indicating the local structures. Compared to the techniques used in natural language processing, we propose a simple and yet efficient approach. For all Transformer modules, coordinates of each input point are firstly mapped to the feature dimension. Then, we subtract the coordinates of the query and key points and use relative positions for encoding. The encoding function is formalized as:

\begin{table*}[t]
    \begin{center}
    \setlength{\tabcolsep}{2.2mm}{
    \renewcommand\arraystretch{1.0}
    \begin{tabular}{c|c||ccc|ccc|ccc}
    \toprule
    \multirow{2}{*}{Method} & \multirow{2}{*}{Modality} & \multicolumn{3}{c|}{Car(IoU=0.7)} & \multicolumn{3}{c|}{Pedestrian (IoU=0.5)} & \multicolumn{3}{c}{Cyclist (IOU=0.5)}\\
    & & Easy & Moderate & Hard & Easy & Moderate & Hard & Easy & Moderate & Hard \\
    \midrule
    PointRCNN \cite{Shi2019PointRCNN} & LiDAR & 85.94 & 75.76 & 68.32 & 49.43 & 41.78 & 38.63 & 73.93 & 59.60 & 53.59 \\
    \midrule
    + Pointformer & LiDAR & \textbf{87.13} & \textbf{77.06} & \textbf{69.25} & \textbf{50.67} & \textbf{42.43} & \textbf{39.60} & \textbf{75.01} & \textbf{59.80} & \textbf{53.99} \\
    \bottomrule
    \end{tabular}}
    \end{center}
    \caption{Performance comparison of PointRCNN with and without Pointformer on KITTI test split by submitting to official test server. The evaluation metric is Average Precision(AP) with IoU threshold 0.7 for car and 0.5 for pedestrian/cyclist.}
    \label{kittitest}
\end{table*}

\begin{table*}[t]
    \begin{center}
    \setlength{\tabcolsep}{1.8mm}{
    \renewcommand\arraystretch{1.0}
    \begin{tabular}{c|c||cccccccccc|c}
    \toprule
    Method & Modality & Car & Ped & Bus & Barrier & TC & Truck & Trailer & Moto & Cons. Veh. & Bicycle & mAP\\
    \midrule
    CBGS \cite{CBGS2019} & LiDAR & 81.1 & 80.1 & 54.9 & 65.7 & 70.9 & \textbf{48.5} & 42.9 & 51.5 & \textbf{10.5} & 22.3 & 52.8\\
    \midrule
    + Pointformer & LiDAR & \textbf{82.3} & \textbf{81.8} & \textbf{55.6} & \textbf{66.0} & \textbf{72.2} & 48.1 & \textbf{43.4} & \textbf{55.0} & 8.6 & \textbf{22.7} & \textbf{53.6}\\
    \bottomrule
    \end{tabular}}
    \end{center}
    \caption{Performance comparison of PointRCNN with and without Pointformer on the nuScenes benchmark.}
    \label{nuscenetest}
\end{table*}

\begin{table}[t]
    \begin{center}
    \setlength{\tabcolsep}{4.2mm}{
    \renewcommand\arraystretch{1.0}
    \begin{tabular}{c|ccc}
    \toprule
    \multirow{2}{*}{Method} & \multicolumn{3}{c}{Car(IoU=0.7)}\\
    & Easy & Moderate & Hard\\
    \midrule
    PointRCNN & 88.88 & 78.63 & 77.38\\
    \midrule
    + Pointformer & \textbf{90.05} & \textbf{79.65} & \textbf{78.89}\\
    \bottomrule
    \end{tabular}}
    \end{center}
    \caption{Performance comparison of PointRCNN with and without Pointformer on the car class of KITTI val split set.}
    \label{kittival}
\end{table}

\begin{table}[t]
    \begin{center}
    \setlength{\tabcolsep}{0.4mm}{
    \renewcommand\arraystretch{1.0}
    \begin{tabular}{c|cc|cc}
    \toprule
    \multirow{2}{*}{RoIs} & \multicolumn{2}{c|}{Recall(IoU=0.5)} & \multicolumn{2}{c}{Recall(IoU=0.7)}\\
    & PointRCNN & +Pointformer & PointRCNN & +Pointformer\\
    \midrule
    10 & 86.66 & \textbf{87.51} & 29.87 & \textbf{35.46}\\
    50 & 96.01 & \textbf{96.52} & 40.28 & \textbf{42.45}\\
    100 & 96.79 & \textbf{96.91} & 74.81 & \textbf{75.82}\\
    200 & \textbf{98.03} & 97.99 & 76.29 & \textbf{76.51}\\
    \bottomrule
    \end{tabular}}
    \end{center}
    \caption{Recall of proposal generation network with different number of RoIs and 3D IoU thresholds for the car class on the val split at moderate difficulty.}
    \label{kittiroi}
\end{table}

\subsection{Computational Cost Reduction}
\label{ccr}
Since Pointformer is a pure attention model based on Transformer blocks, it suffers from extremely heavy computational overhead. Applying a conventional Transformer to a point cloud with $n$ points consumes $O(n^2)$ time and memory, leading to much more training cost. 

Some recent advances in efficient Transformers have mitigated this issue \cite{kitaev2020reformer, Katharopoulos2020TransformersAR, wang2020linformer, choromanski2020performer, xie2020mlcvnet}, among which Linformer \cite{wang2020linformer} reduces the complexity to $O(n)$ by low-rank factorization of the original attention. Under the hypothesis that the self attention mechanism is low rank, i.e. the rank of the $n\times{}n$ attention matrix
\begin{equation}
    A = \text{softmax}\left(\cfrac{QK^\top}{\sqrt{d_k}}\right),
\end{equation}
is much smaller than $n$, Linformer projects the $n$-dimension keys and values to the ones with lower dimension $k\ll{}n$, and $k$ is closer to the rank of $A$. Therefore, the $i$-th head in the projected multi-head self-attention is
\begin{equation}
    \text{head}_i=\text{softmax}\left(\cfrac{Q(E_iK)^\top}{\sqrt{d_k}}\right)F_iV,
\end{equation}
where $E_i,F_i\in\mathbb{R}^{k\times{}n}$ are the projection matrices, which reduces the complexity from $O(n^2)$ to $O(kn)$. 

Compared with the Taylor expansion approximation technique used in MLCVNet \cite{xie2020mlcvnet}, Linformer is easier to implement in out method. We thus adopt it to replace the Transformer layers in the vanilla Pointformer. Practically, we map the number of points $n$ to $k=\frac{n}{r}$, where $r$ is a factor controlling the number of projected dimensions. We apply this mapping in Local Transformer, Global Transformer and Local-Global Transformer blocks. By setting an appropriate factor $r$ for each block, there would be a significant boost in both time and space consumption with little performance decay.

\begin{table*}[t]
    \begin{center}
    \setlength{\tabcolsep}{2.4mm}{
    \renewcommand\arraystretch{1.0}
        \begin{tabular}{c||cccccccccc|c}
		\toprule
		Method & bathtub & bed & bookshelf & chair & desk & dresser & nightstand & sofa & table & toilet & mAP \\
		\midrule
		VoteNet \cite{qi2019deep} & 74.4 & 83.0 & 28.8 & 75.3 & 22.0 & 29.8 & 62.2 & 64.0 & 47.3 & 90.1 & 57.7 \\
		VoteNet* & 75.5 & \textbf{85.6} & \textbf{32.0} & \textbf{77.4} & 24.8 & 27.9 & 58.6 & \textbf{67.4} & 51.1 & 90.5 & 59.1 \\
		\midrule
		+ Pointformer & \textbf{80.1} & 84.3 & \textbf{32.0} & 76.2 & \textbf{27.0} & \textbf{37.4} & \textbf{64.0} & 64.9 & \textbf{51.5} & \textbf{92.2} & \textbf{61.1} \\
		\bottomrule
	\end{tabular}
	}
    \end{center}
	\caption{Perfomance comparison of VoteNet with and without Pointformer on \textbf{SUN RGB-D} validation dataset. The evaluation metric is Average Precision with \textbf{0.25 IoU threshold}.* denotes the model implemented in MMDetection3D ~\cite{mmdetection3d}.}
	\label{sunrbgddataset}
\end{table*}

\begin{table*}[t]
    \begin{center}
    \setlength{\tabcolsep}{0.65mm}{
    \renewcommand\arraystretch{1.0}
    \begin{tabular}{c||cccccccccccccccccc|c}
		\toprule
		Method & cab & bed & chair & sofa & table & door & wind & bkshf & pic & cntr & desk & curt & fridg & showr & toil & sink & bath & ofurn & mAP \\
		\midrule
		VoteNet \cite{qi2019deep} & 36.3 & 87.9 & 88.7 & \textbf{89.6} & 58.8 & 47.3 & 38.1 & 44.6 & 7.8 & 56.1 & 71.7 & 47.2 & 45.4 & 57.1 & 94.9 & 54.7 & 92.1 & 37.2 & 58.6 \\
		VoteNet* & \textbf{47.7} & \textbf{88.7} & 89.5 & 89.3 & 62.1 & 54.1 & 40.8 & 54.3 & 12.0 & \textbf{63.9} & \textbf{69.4} & 52.0 & \textbf{52.5} & \textbf{73.3} & 95.9 & 52.0 & \textbf{95.1} & 42.4 & 62.9 \\
		\midrule
		+ Pointformer & 46.7 & 88.4 & \textbf{90.5} & 88.7 & \textbf{65.7} & \textbf{55.0} & \textbf{47.7} & \textbf{55.8} & \textbf{18.0} & 63.8 & 69.1 & \textbf{55.4} & 48.5 & 66.2 & \textbf{98.9} & \textbf{61.5} & 86.7 & \textbf{47.4} & \textbf{64.1} \\
		\bottomrule
	\end{tabular}
	}
    \end{center}
    \caption{Performance comparison of VoteNet with and without Pointformer on \textbf{ScanNetV2} validation dataset. The evaluation metric is Average Precision with \textbf{0.25 IoU threshold}.* denotes the model implemented in MMDetection3D ~\cite{mmdetection3d}.}
    \label{scannet}
\end{table*}

\section{Experimental Results}
In this section, we use \textit{Pointformer} as the backbone for state-of-the-art object detection models and conduct experiments on several indoor and outdoor benchmarks. In Sec.~\ref{ES}, we introduce the implementation details of the experiments. In Sec.~\ref{od} and Sec.~\ref{id}, we show the comparison results on indoor and outdoor datasets respectively. In Sec.~\ref{AS}, we conduct extensive ablation studies to analyze our proposed Pointformer model. Finally, we show qualitative results in Sec.~\ref{QRD}. More analysis and visualizations are provided in the appendix.

\subsection{Experimental Setup}
\label{ES}
\noindent 
\textbf{Datasets.} We adopt SUN RGB-D \cite{song2015sun} and ScanNet V2 \cite{dai2017scannet} for indoor 3D detection benchmark. SUN RGB-D has 5K training images annotated with oriented 3D bounding boxes for 37 object categories and ScanNet V2 has 1513 labeled scenes with 40 semantic classes. We follow the same setting in VoteNet \cite{qi2019deep} and report performance on the 10 classes on SUN RGB-D and 18 classes on ScanNet V2. 
For outdoor datasets, we choose KITTI \cite{Geiger2012KITTI} and nuScenes \cite{Caesar2020nuScenesAM} for evaluation. KITTI contains 7,481 training samples and 7,518 test samples for autonomous driving. NuScenes contains 1k different scenes with 40K key frames, which has 23 categories and 8 attributes. We follow the evaluation protocol proposed along with the datasets.

\noindent 
\textbf{Experimental setups.} We use the Pointformer as the backbone for three 3D detection models, including VoteNet \cite{qi2019deep}, PointRCNN \cite{Shi2019PointRCNN} and CBGS \cite{CBGS2019}. VoteNet is a point-based approach for indoor datasets, while PointRCNN and CBGS are adopted for outdoor datasets. PointRCNN is a classic approach for autonomous driving detection and CBGS is the champion of nuScenes 3D detection Challenge held in CVPR 2019.
For a fair comparison, we adopt the same detection head, number of points for each resolution, hyperparameters and training configurations as baseline models. 

\subsection{Outdoor Datasets}
\label{od}
\noindent 
\textbf{KITTI.}
We first evaluate our method comparing with PointRCNN on KITTI's 3D detection benchmark. PointRCNN uses PointNet++ as its backbone with four set abstraction layers. Similarly, we adopt the same architecture, while switching the set abstraction layer in PointNet++ with the proposed Transformer block. The comparison results on the KITTI test server are shown in Table \ref{kittitest}.

For the car category, we also report the performance of 3D detection results on the val split as shown in Table \ref{kittival}. As we can observe, by adopting Pointformer, our model achieves consistent improvements comparing to the original PointRCNN. Especially in the hard difficulty, our method shows the most promising result with 1.5\% AP improvement. We believe the better performance on hard objects is attributed to the higher expressiveness of local Transformer module. For hard objects which are often small or occluded, GT captures context-dependent region features, which contributes to the bounding box regression and classification.

Additionally, we evaluate the performance of proposal generation network by calculating the recall of 3D bounding box with various number of proposals and 3D IoU threshold. As shown in Table \ref{kittiroi}, our backbone module significantly enhances the performance of proposal generation network under almost all the settings. Analyzing the figures vertically, we observe that our backbone shows better performance when the number of RoIs are relatively small. As stated in Sec.3, the GT and LGT help to capture context-aware representations and models the relations among different objects (proposals). This provides additional references for locating and reasoning the bounding boxes. Therefore, despite the lack of RoIs, we can still improve the performance of the proposal generation module and achieve higher recall.

\noindent 
\textbf{NuScenes.} We also validate the effectiveness of Pointformer on the nuScenes dataset, which greatly extends KITTI in dataset size, number of object categories and number of annotated objects. Furthermore, nuScenes suffers from severe class imbalance issues, making the detection task more difficult and challenging. In this part, we adopt CBGS, the champion of nuScenes 3D detection Challenge held in CVPR 2019, as the baseline model and show the comparison results when replacing the backbone with Pointformer. We summarize the results in Table \ref{nuscenetest}. As we can observe, by utilizing Pointformer as the backbone, our model achieves 0.8 higher mAP than baseline. For 8 of 10 classes, our model shows better performance, which demonstrates the effectiveness of Pointformer on larger and more challenging datasets.

\subsection{Indoor Datasets}
\label{id}

\noindent
We evaluate our Pointformer accompanied by VoteNet \cite{qi2019deep} on SUN RGB-D and ScanNet V2. We follow the same hyperparameters on the backbone structure as VoteNet. Followed by the Pointformer blocks, two feature propagation(FP) modules proposed in PointNet++ \cite{qi2017pointnet++} serve as upsamplers to increase the resolution for the subsequent detection heads.

\noindent
\textbf{SUN RGB-D.} We report the average precision(AP) over 10 common classes in SUN RGB-D, as shown in Table \ref{sunrbgddataset}. Compared with the PointNet++ \cite{qi2017pointnet++} in VoteNet \cite{qi2019deep}, our Pointformer provides a significant boost with 2$\%$ mAP over the implementation in MMDetection3D \cite{mmdetection3d}. On some categories with large and complex objects like dresser or bathtub, Pointformer shows its splendid capability on extracting non-local information by a sharp increase over 5$\%$ AP, which we attribute to the GT module in Pointformer.

\noindent
\textbf{ScanNet V2.} We report the average precision(AP) over 18 classes in ScanNet V2, as shown in Table \ref{scannet}. Compared with VoteNet, Pointformer outperforms its original version by 1.2$\%$ mAP with MMDetection3D. 

\begin{table}[t]
    \begin{center}
    \setlength{\tabcolsep}{1.8mm}{
    \renewcommand\arraystretch{1.0}
    \begin{tabular}{c|cccc|ccc}
    \toprule
    \multirow{2}{*}{} & \multirow{2}{*}{LT} & \multirow{2}{*}{GT} & \multirow{2}{*}{LGT} & \multirow{2}{*}{CoRe} & \multicolumn{3}{c}{Car (IoU=0.7)}\\
    &&&&& Easy & Moderate & Hard\\
    \midrule
    1 & - & - & - & - & 88.88 & 78.63 & 77.38 \\
    2 & \checkmark & - & - & - & 89.46 & 78.91 & 77.65\\
    3 & \checkmark & - & - & \checkmark & 89.76 & 79.24 & 78.43\\
    4 & \checkmark & \checkmark & - & - & 89.68 & 79.22 & 78.52\\
    5 & \checkmark & \checkmark & \checkmark & - & 89.82 & 79.34 & 78.62\\
    6 & \checkmark & \checkmark & \checkmark & \checkmark & \textbf{90.05} & \textbf{79.65} & \textbf{78.89}\\
    \bottomrule
    \end{tabular}}
    \end{center}
    \caption{Effects of each component on the val split of KITTI. CoRe represents the coordinates refinement module.}
    \label{ablation1}
\end{table}

\begin{table}[t]
    \begin{center}
    \setlength{\tabcolsep}{2.0mm}{
    \renewcommand\arraystretch{1.0}
    \begin{tabular}{c|c|ccc}
    \toprule
    \multirow{2}{*}{} & \multirow{2}{*}{Positional Encoding} & \multicolumn{3}{c}{Car (IoU=0.7)}\\
    && Easy & Moderate & Hard\\
    \midrule
    1 & - & 85.42 & 75.67 & 72.34 \\
    2 & \checkmark & \textbf{90.05} & \textbf{79.65} & \textbf{78.89}\\
    \bottomrule
    \end{tabular}}
    \end{center}
    \caption{Effects of positional encoding on the val split of KITTI.}
    \label{ablation2}
\end{table}

\begin{figure*}
\begin{center}
\begin{tabular}{ccc|ccc}
    Image of the scene & Predictions & Ground truth & Image of the scene & Predictions & Ground truth \\
    \includegraphics[width=0.14\linewidth]{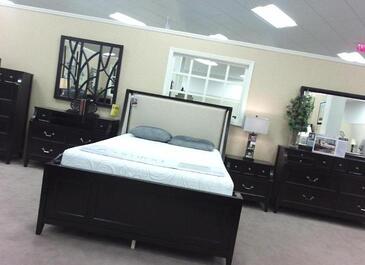} & \includegraphics[width=0.14\linewidth]{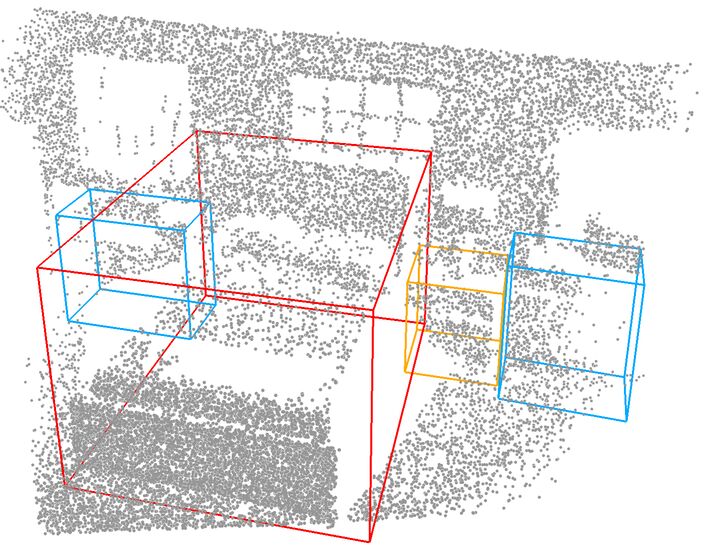} & \includegraphics[width=0.14\linewidth]{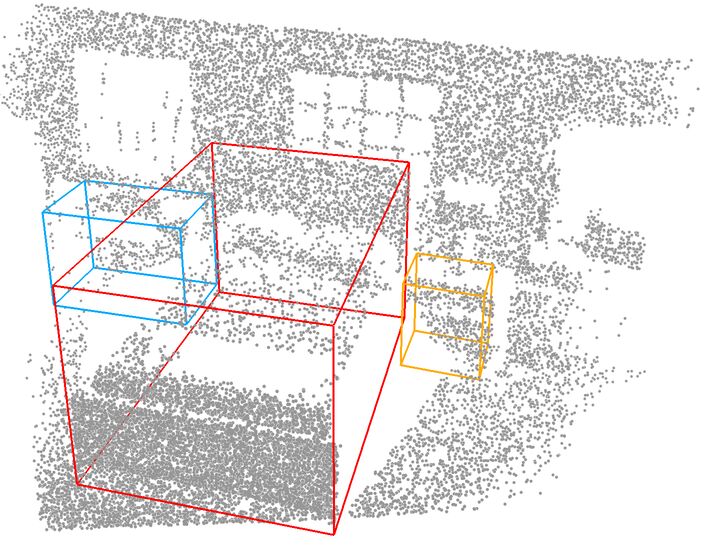} &
    \includegraphics[width=0.14\linewidth]{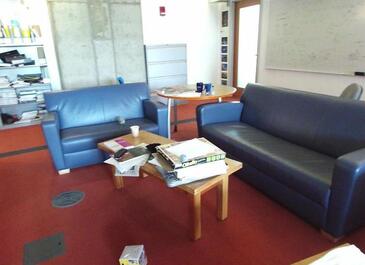} & \includegraphics[width=0.14\linewidth]{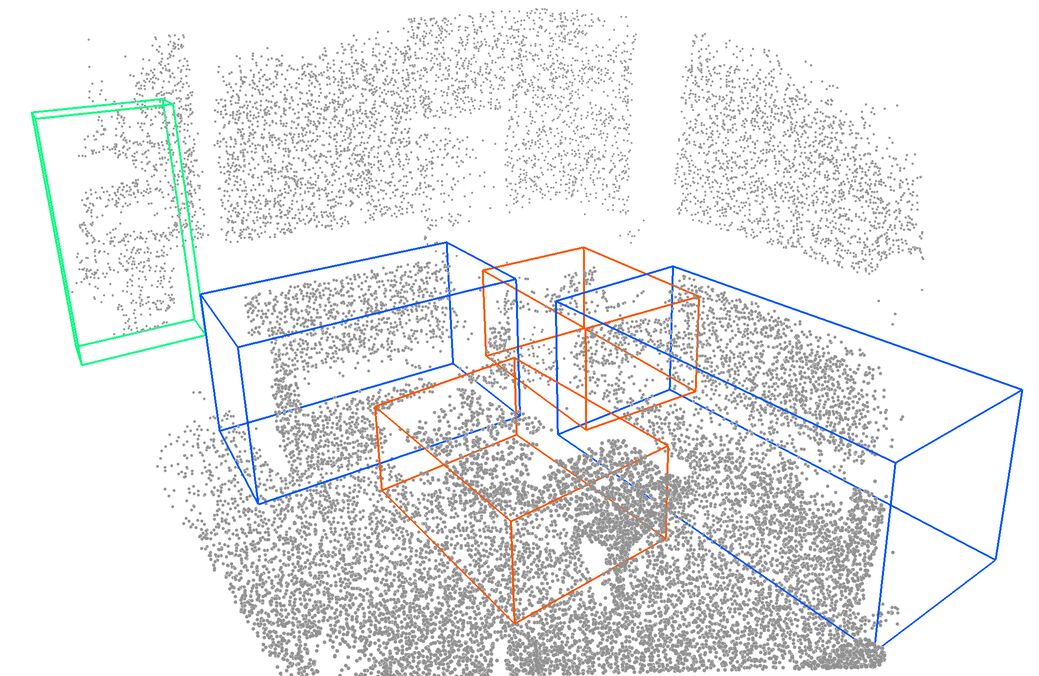} & \includegraphics[width=0.14\linewidth]{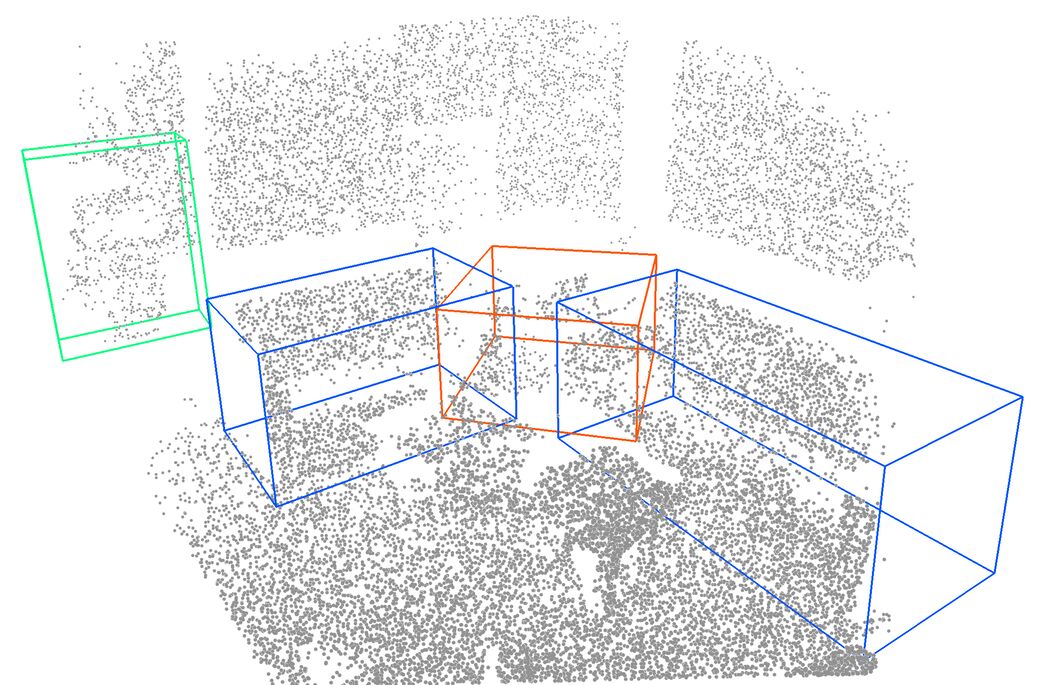} \\
\end{tabular}
\end{center}
\caption{\textbf{Qualitative results of 3D object detection on SUN RGB-D.} From left to right: Originial scene image, our model's prediction, and annotated ground truth boxes.}
\label{fig:vissunrgbd}
\end{figure*}

\begin{figure}
    \begin{center}
    \begin{tabular}{ccc}
        Image of the scene &
        Ground truth &
        Overall attention \\
        \includegraphics[width=0.28\linewidth]{sun03img.jpg} &
        \includegraphics[width=0.28\linewidth]{sun03gt.jpg} & 
        \includegraphics[width=0.28\linewidth]{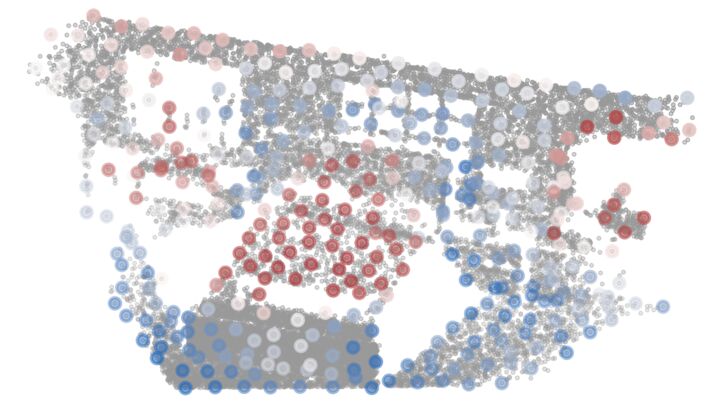}\\
    \end{tabular}
    \begin{tabular}{ccc}
         Top-50 attention &
        Top-100 attention &
        Top-200 attention \\
         \includegraphics[width=0.28\linewidth]{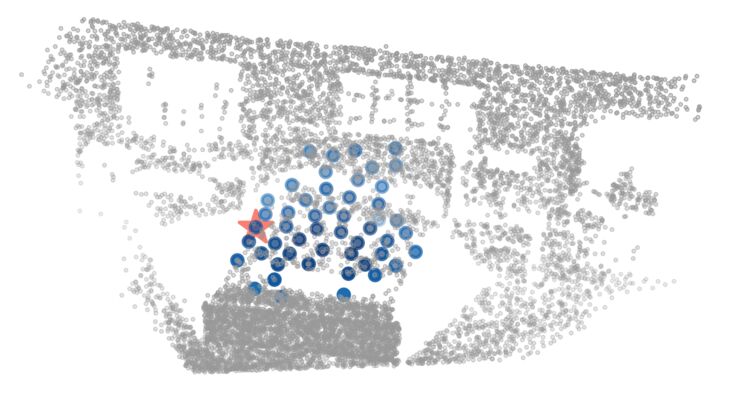} &
        \includegraphics[width=0.28\linewidth]{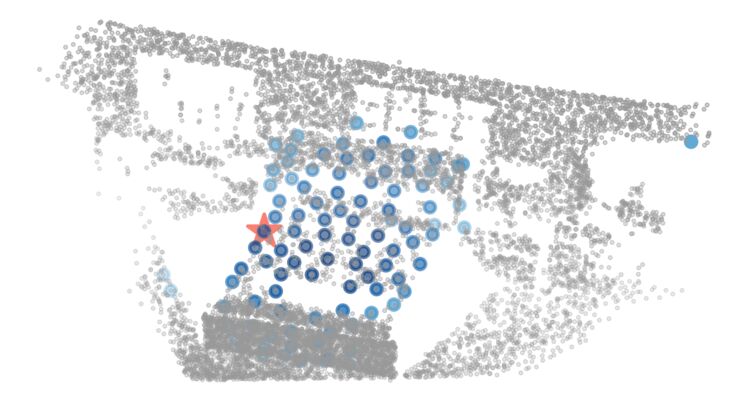} &
        \includegraphics[width=0.28\linewidth]{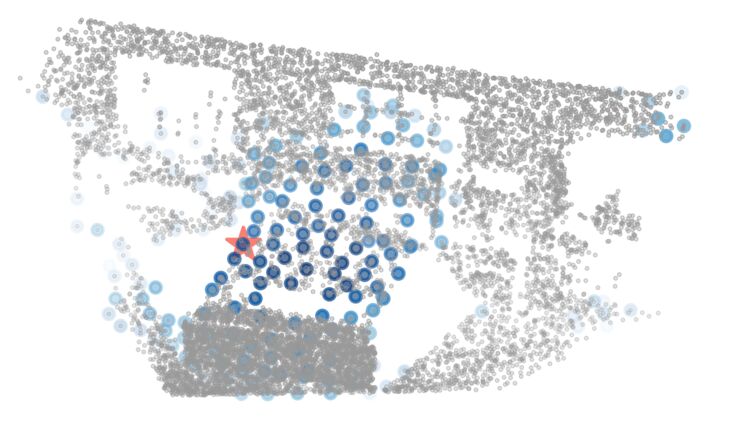} \\
    \end{tabular}
    \caption{\textbf{Visualization results of the attention maps.} In top-k attention, darker color indicates larger attention weight, in overall attention red indicates large value.}
    \end{center}
    \label{fig:attvis}
\end{figure}

\subsection{Ablation Study}
\label{AS}
In this section, we conduct extensive ablation experiments to analyze the effectiveness of different components of Pointformer. All experiments are trained on the train split with PointRCNN detection head and evaluated on the val split with the car class.

\noindent 
\textbf{Effects of each component.} We validate the effectiveness of each Transformer component and the coordinate refinement module, and summarized the results in Table \ref{ablation1}. The first row corresponds to the PointRCNN baseline and the last row is the full Pointformer model. By comparing the first row and second row, we can observe that easy objects benefit more from the local Transformer with ~$0.6$ AP improvement. By comparing the second row and fourth row, we can see that global Transformer is more suitable for hard objects with ~$0.9$ AP improvement. This observation is consistent with our analysis in Sec.~\ref{od}. As for Local-Global Transformer and coordinate refinement, the improvement is similar under three difficulty settings.

\noindent
\textbf{Positional Encoding.} 
Playing a critical role in Transformer, position encoding can have huge impact on the learned representation. As we have shown in Table \ref{ablation2}, we compare the performance of Pointformer without positional encoding and with two approaches to position encoding (adding or concatenating positional encoding with the attention map). We can observe that Pointformer without positional encoding suffers from a huge performance drop, as the coordinates of points can capture the local geometric information. 

\subsection{Qualitative Results and Discussion}
\label{QRD}

\noindent
\textbf{Qualitative results on SUN RGB-D.} Figure \ref{fig:vissunrgbd} shows representative examples of detection results on SUN RGB-D with VoteNet + Pointformer. As we can observe, our model achieves robust results despite the challenges of clutter and scanning artifacts. Additionally, our model can even recognize the missing objects in the ground truth. For instance, the dresser in the left scene is only partially observed by the sensor. However, our model can still generate precise proposals for the object with proper bounding box sizes. Similar results are shown in the right scene, where the table in the front suffers from clutter because of the books on it.

\noindent
\textbf{Inspecting Pointformer with attention maps.} To validate how modules in Pointformer affect learned point features, we visualize the attention maps from the GT module of the second last Pointformer block. We show the attention of the particular points in Figure 5. The second row shows the 50, 100, 200 points with highest attention values towards the points marked with star. We can observe that Pointformer first focuses on the local region of the same object, then spread the attention to other regions, and finally attends points from other objects globally. The overall attention map shows the average attention weights of all the points in the scene, indicating that our model mostly focuses on points on the objects. These visualization results show that Pointformer can capture local and global dependencies, and enhance message passing on both object and scene levels.
\section{Conclusion}

This paper introduces \textit{Pointformer}, a highly effective feature learning backbone for 3D point clouds that is permutation invariant to points in the input and learns local and global context-aware representations. We apply Pointformer as the drop-in replacement backbone for state-of-the-art 3D object detectors and show significant performance improvements on several benchmarks including both indoor and outdoor datasets.

Comparing to classification and segmentation tasks including part-segmentation and semantic segmentation in prior work, 3D object detection typically involves more points (4$\times$ - 16$\times$) in a scene, which makes it harder for Transformer-based models. For future work, we would like to explore extensions to these two tasks and other 3D tasks such as shape completion, normal estimation, etc.

\vspace{-0.1in}
\section*{Acknowledgments}
\vskip -0.05in

This work is supported in part by the National Science and Technology Major Project of the Ministry of Science and Technology of China under Grants 2018AAA0100701, the National Natural Science Foundation of China under Grants 61906106 and 62022048, the Institute for Guo Qiang of Tsinghua University and Beijing Academy of Artificial Intelligence.

{\small
\bibliographystyle{ieee_fullname}
\bibliography{egbib}
}
\begin{appendix}

\clearpage
\section*{Appendix}
\section{Architectures and Implementation Details}

In this section, we discuss each component in our \textit{Pointformer} models for indoor and outdoor settings in detail.

\noindent
\textbf{Indoor Datasets.} First, the Local Transformer(LT) block is composed of a sequence of sampling and grouping operations, followed by a shared positional encoding layer and two self-attention transformer layers, with a linear shared Feed-Forward Network(FFN) in the end. As shown in Table \ref{model_arch_indoor}, we use the same sampling and grouping parameters, and feature dimensions as those of PointNet++ \cite{qi2017pointnet++}, the backbone in VoteNet \cite{qi2019deep} and H3DNet \cite{zhang2020h3dnet}.

Second, the Local-Global Transformer(LGT) and the Global Transformer(LT) have fewer hyper-parameters than LT, where we adopt two self-attention layers in GT and one cross-attention layer in LGT for each Pointformer block. Since their massive attention computation may lead to overfitting, we apply dropout with the dropping probability 0.4 on SUN RGB-D \cite{song2015sun} and 0.2 on ScanNetV2 \cite{dai2017scannet}. As for the number of heads in multi-head attention, we set it to 8 on ScanNetV2 and 4 on SUN RGB-D. In our experiments, we found that noisy backgrounds in indoor datasets affect the LGT performance, by reducing 1$\sim$2$\%$mAP. So we report the results on indoor datasets without the LGT module.

\begin{table}[ht]
    \begin{center}
    \setlength{\tabcolsep}{1mm}{
    \begin{tabular}{c||c|c|c|c|c|c|c}
		\toprule
		\#block & $N_\text{in}$ & $N_\text{out}$ & radius & samples & $C_\text{in}$ & $C_\text{med}$ & $C_\text{out}$ \\
		\midrule
		1 & $N_\text{point}$ & 2048 & 0.2 & 64 & 64 & 64 & 128 \\
		2 & 2048 & 1024 & 0.4 & 32 & 128 & 128 & 256 \\
		3 & 1024 & 512 & 0.8 & 16 & 256 & 256 & 512 \\
		4 & 512 & 256 & 1.2 & 16 & 512 & 512 & 512 \\
		\bottomrule
	\end{tabular}}
    \end{center}
    \caption{\textbf{Model Architecture details on indoor datasets.} $N_\text{in}$ denotes the number of input points to this Pointformer block, and $N_\text{out}$ is the number of sampled output points of the block. Radius and samples are hyper-parameters of ball query operation to gather points in a neighborhood in LT. $C_\text{in}$, $C_\text{med}$ and $C_\text{out}$ denote the dimensions of features in LT, LGT and GT respectively. $N_\text{point}$ is the scale of original point clouds in the dataset, 20,000 for SUN RGB-D and 40,000 for ScannNetV2.}
    \label{model_arch_indoor}
\end{table}

Finally, we implement our indoor models on the top of MMDetection3D, an open source toolbox 3D object detection. We follow the same hyper-parameters and data augmentation techniques as those of VoteNet. To train a \textit{Pointformer} on SUN RGB-D, we use the AdamW \cite{adam,adamw} optimizer with an initial learning rate of 3e-4 and weight decay factor of 0.05, and decay the learning rate by 0.3 at epoch 24 and 32 during the training of a total of 36 epochs. And for ScanNet, we use AdamW optimizer with 0.002 learning rate and set 0.1 weight decay. We decay the learning rate by 0.3 at epoch 32 and 40 during the training of 48 epochs.

\noindent
\textbf{Outdoor Datasets.}
We adopt the same structure of Transformer blocks as that for indoor datasets. Two self-attention layers with FFN are adopted in LT and GT, while only one cross-attention layer is utilized in LGT block. The number of heads are set to 8 for both KITTI \cite{Geiger2012KITTI} and nuScenes \cite{Caesar2020nuScenesAM}.

\begin{table}[ht]
    \begin{center}
    \setlength{\tabcolsep}{1mm}{
    \begin{tabular}{c||c|c|c|c|c|c|c}
		\toprule
		\#block & $N_\text{in}$ & $N_\text{out}$ & radius & samples & $C_\text{in}$ & $C_\text{med}$ & $C_\text{out}$ \\
		\midrule
		1 & 16384 & 4096 & 0.1 & 64 & 64 & 64 & 128 \\
		2 & 4096 & 1024 & 0.5 & 32 & 128 & 128 & 256 \\
		3 & 1024 & 256 & 1.0 & 16 & 256 & 256 & 512 \\
		4 & 256 & 64 & 2.0 & 16 & 512 & 512 & 512 \\
		\bottomrule
	\end{tabular}
	}
    \end{center}
    \caption{\textbf{Model Architecture details on KITTI datasets.}}
    \label{model_arch_kitti}
\end{table}

We implement our outdoor models on top of OpenPCDet \cite{openpcdet2020}, an open source toolbox for LiDAR-based 3D object detection. We follow the same hyper-parameters as that of PointRCNN, including data augmentation, post-processing, etc. To train a Pointformer on KITTI, we use the Adam optimizer with an initial learning rate of 5e-3 and weight decay of 0.01.

\begin{table*}[!htbp]
    \begin{center}
    \setlength{\tabcolsep}{0.7mm}{
    \begin{tabular}{c||cccccccccccccccccc|c}
		\toprule
		Method & cab & bed & chair & sofa & table & door & wind & bkshf & pic & cntr & desk & curt & fridg & showr & toil & sink & bath & ofurn & mAP \\
		\midrule\midrule
		VoteNet \cite{qi2019deep} & 8.1 & 76.1 & 67.2 & 68.8 & 42.4 & 15.3 & 6.4 & 28.0 & 1.3 & 9.5 & 37.5 & 11.6 & 27.8 & 10.0 & 86.5 & 16.8 & 78.9 & 11.7 & 33.5 \\
		VoteNet* & 14.6 & 77.9 & 73.1 & \textbf{80.5} & 46.5 & \textbf{25.1} & \textbf{16.0} & 41.8 & \textbf{2.5} & 22.3 & 33.3 & 25.0 & 31.0 & 17.6 & \textbf{87.8} & 23.0 & 81.6 & 18.7 & 39.9 \\
		\midrule
		+Pointformer & \textbf{19.0} & \textbf{80.0} & \textbf{75.3} & 69.0 & \textbf{50.5} & 24.3 & 15.0 & \textbf{41.9} & 1.5 & \textbf{26.9} & \textbf{45.1} & \textbf{30.3} & \textbf{41.9} & \textbf{25.3} & 75.9 & \textbf{35.5} & \textbf{82.9} & \textbf{26.0} & \textbf{42.6} \\
		\bottomrule
	\end{tabular}
	}
    \end{center}
    \caption{Performance comparison of VoteNet with and without Pointformer on \textbf{ScanNetV2} validation dataset. The evaluation metric is Average Precision with \textbf{0.5 IoU threshold}.* denotes the model implemented in MMDetection3D \cite{mmdetection3d}.}
    \label{scan05}
\end{table*}

\begin{table*}[!htbp]
    \begin{center}
    \setlength{\tabcolsep}{2.4mm}{
        \begin{tabular}{c||cccccccccc|c}
		\toprule
		Method & bathtub & bed & bookshelf & chair & desk & dresser & nightstand & sofa & table & toilet & mAP \\
		\midrule\midrule
		VoteNet \cite{qi2019deep} & \textbf{49.9} & 47.3 & 4.6 & 54.1 & 5.2 & 13.6 & 35.0 & 41.4 & 19.7 & 58.6 & 32.9 \\
		VoteNet* & 43.5 & 55.9 & \textbf{7.2} & \textbf{56.5} & \textbf{5.7} & 12.6 & 39.7 & 50.1 & 20.7 & \textbf{66.3} & 35.8 \\
		\midrule
		+Pointformer & 42.5 & \textbf{59.0} & 6.3 & 54.2 & 5.4 & \textbf{20.5} & \textbf{43.3} & \textbf{51.0} & \textbf{22.4} & 61.2 & \textbf{36.6} \\
		\bottomrule
	\end{tabular}
	}
    \end{center}
	\caption{Performance comparison of VoteNet with and without Pointformer on \textbf{SUN RGB-D} validation dataset. The evaluation metric is Average Precision with \textbf{0.5 IoU threshold}.* denotes the model implemented in MMDetection3D \cite{mmdetection3d}.}
	\label{rgbd05}
\end{table*}

\begin{table}
    \begin{center}
    \setlength{\tabcolsep}{1mm}{
        \begin{tabular}{c||cc}
		\toprule
		Method &  mAP@0.25 & mAP@0.5 \\
		\midrule
		H3DNet* - 1 tower & 64.1 & 44.2 \\
		\midrule
		+Pointformer & \textbf{64.4} & \textbf{44.4} \\
		\bottomrule
	\end{tabular}
	}
    \end{center}
	\caption{Performance comparison of H3DNet \cite{zhang2020h3dnet} with and without Pointformer on \textbf{ScanNet V2} validation dataset. For fair comparison we use single backbone instead of multiple backbones. * denotes the model implemented in MMDetection3D \cite{mmdetection3d}.}
	\label{h3d1t}
\end{table}

\begin{table}[t]
    \begin{center}
    \setlength{\tabcolsep}{1.4mm}{
    \begin{tabular}{c|c|ccc}
    \toprule
    Method & \multirow{2}{*}{Params} & \multicolumn{3}{c}{Car (IoU=0.7)}\\
    (PointRCNN+) && Easy & Moderate & Hard\\
    \midrule
    PointNet++(default) & 4.04M & 88.88 & 78.63 & 77.38 \\
    Pointformer(small) & 4.12M & \textbf{89.35} & \textbf{79.01} & \textbf{78.34}\\
    \midrule
    PointNet++(large) & 6.24M & 89.01 & 78.82 & 77.67\\
    Pointformer(default) & 6.06M & \textbf{90.05} & \textbf{79.65} & \textbf{78.89}\\
    \bottomrule
    \end{tabular}}
    \end{center}
    \caption{Comparison of PointNet++ and Pointformer with similar parameters on the val split of KITTI.}
    \label{ablation3}
\end{table}

\begin{table}[t]
    \begin{center}
    \setlength{\tabcolsep}{1.0mm}{
    \begin{tabular}{c|c|ccc}
    \toprule
    \multirow{2}{*}{Method} & \multirow{2}{*}{Latency} & \multicolumn{3}{c}{Car (IoU=0.7)}\\
    && Easy & Moderate & Hard\\
    \midrule
    Poinftormer+Linformer & 0.22 & 89.94 & 79.63 & 78.85\\
    Pointformer & 0.25 & \textbf{90.05} & \textbf{79.65} & \textbf{78.89}\\
    \bottomrule
    \end{tabular}}
    \end{center}
    \caption{Performance of Pointformer with and withour the Linformer technique on the val split of KITTI.}
    \label{ablation4}
\end{table}

\section{More Quantitative Results}
In this section, we provide more results and analysis on SUN RGB-D and ScanNetV2 as shown in Table \ref{scan05}\&\ref{rgbd05}. With 0.5 IoU threshold, our proposed Pointformer achieves consistent improvements on both dataset. In Table \ref{h3d1t}, we use the one tower version H3DNet \cite{zhang2020h3dnet} as baseline, showing our method can work well with the recent advanced model.

\section{More Ablation Studies}
\noindent
\textbf{Parameter Efficiency.} To further validate the effectiveness of Pointformer, we conduct experiments and compare the backbones with similar model parameters. We reduce the Transformer layers adopted in each block and refer the model as Pointformer(small). Similarly, we increase the FFN layers in PointNet++ and refer the model as PointNet++(large). As we have shown in Table \ref{ablation3}, Pointformer achieves better results under both parameter budgets. Although our model suffers from a performance reduction when using fewer Transformer layers, we are still 0.5\%  to  1\% AP higher for all difficulty levels. Additionally, PointNet++ shows little improvement with larger feature dimensions. By comparison, Pointformer can adapt to deeper models and use learning parameters more efficiently.

\noindent
\textbf{Computational Cost Reduction.} As stated in Section \ref{ccr}, Transformer-based modules suffer from heavy computational cost and memory consumption. Therefore, we adopt the Linformer technique to improve model efficiency. The results are shown in Table \ref{ablation4} and we can observe that inference latency is decreased with little drop in performance.

\section{More Qualitative Results}
\begin{figure*}
\begin{center}
\begin{tabular}{cccccc}
    Image of the scene & Ground truth & Top-50 attention & Top-100 attention & Top-200 attention & Overall attention \\
    \includegraphics[width=0.14\linewidth]{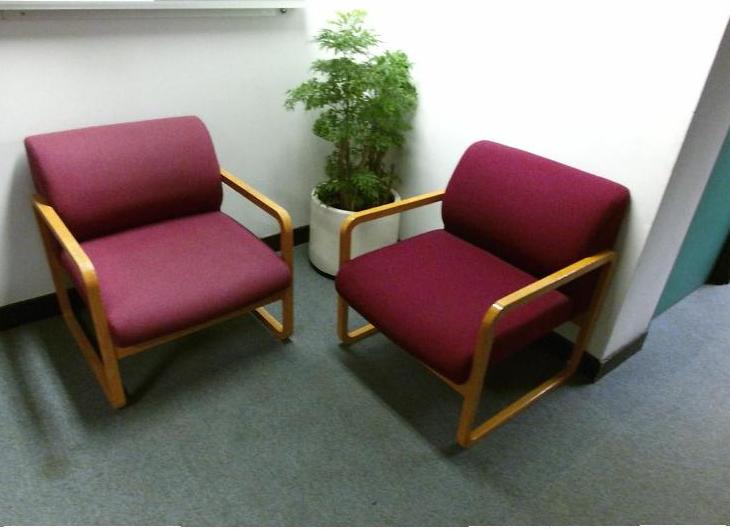} & \includegraphics[width=0.14\linewidth]{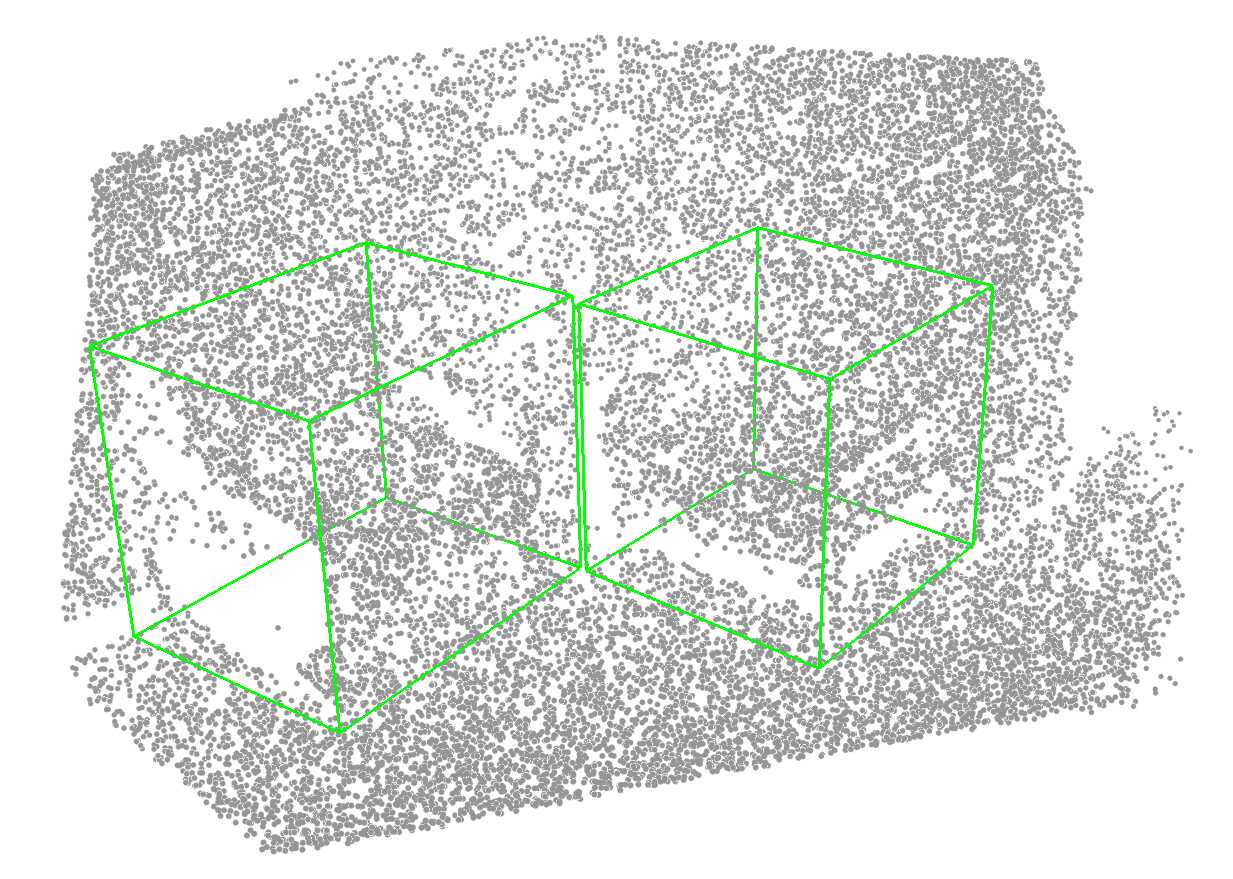} & \includegraphics[width=0.14\linewidth]{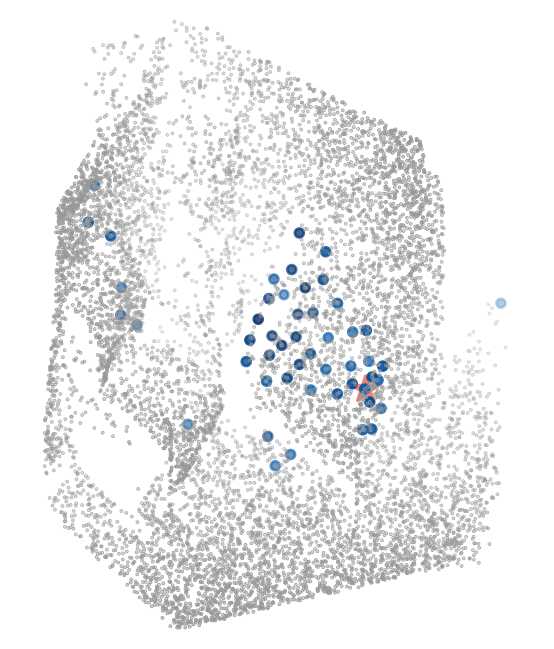} &
    \includegraphics[width=0.14\linewidth]{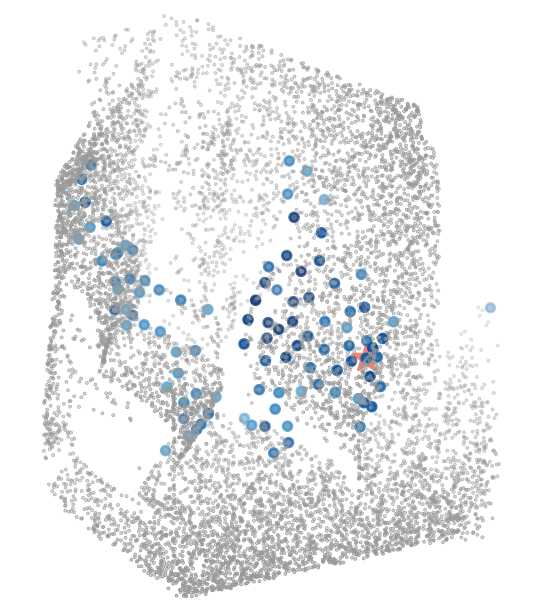} & \includegraphics[width=0.14\linewidth]{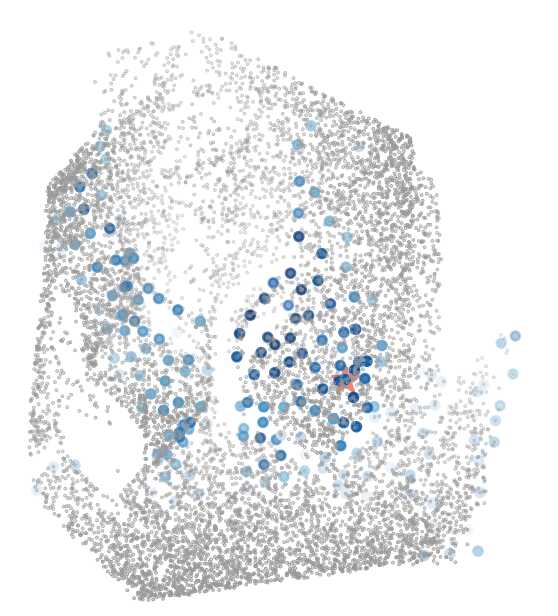} & \includegraphics[width=0.14\linewidth]{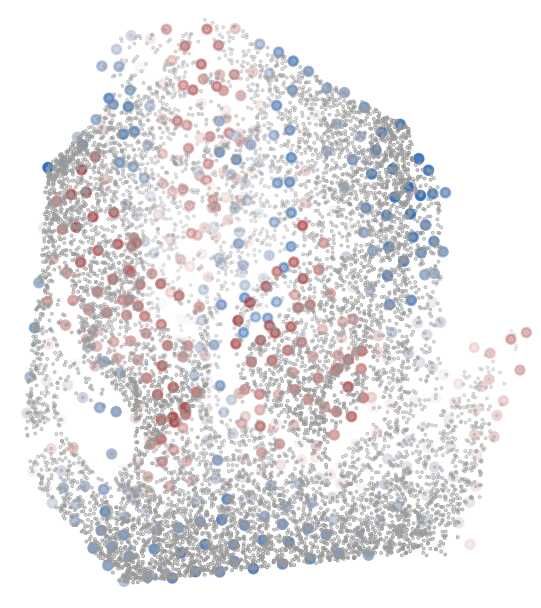} \\
    \includegraphics[width=0.14\linewidth]{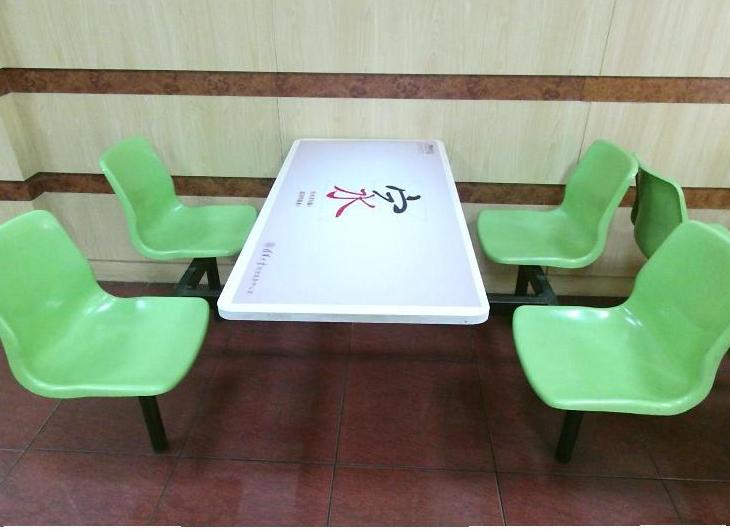} & \includegraphics[width=0.14\linewidth]{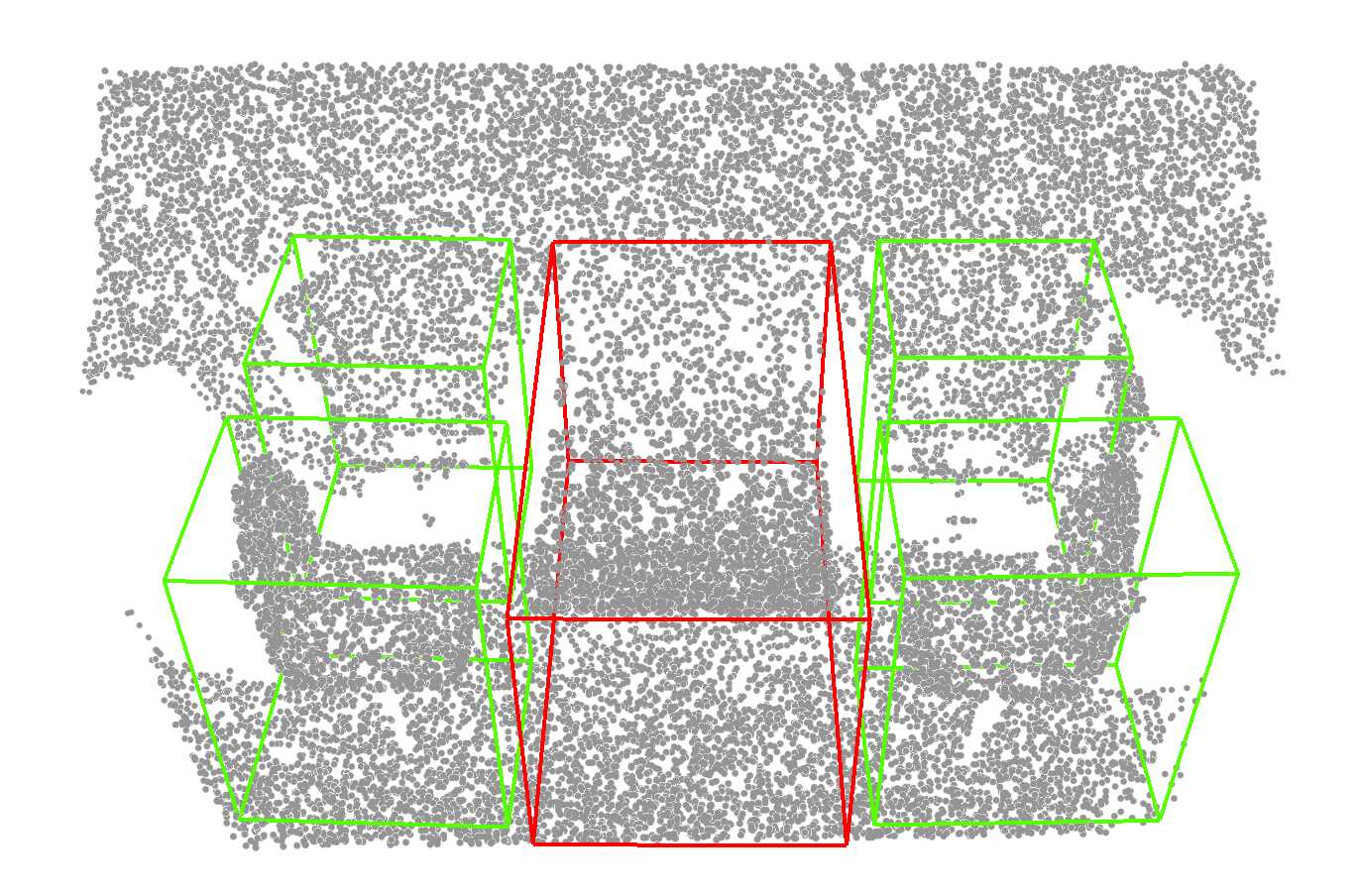} & \includegraphics[width=0.14\linewidth]{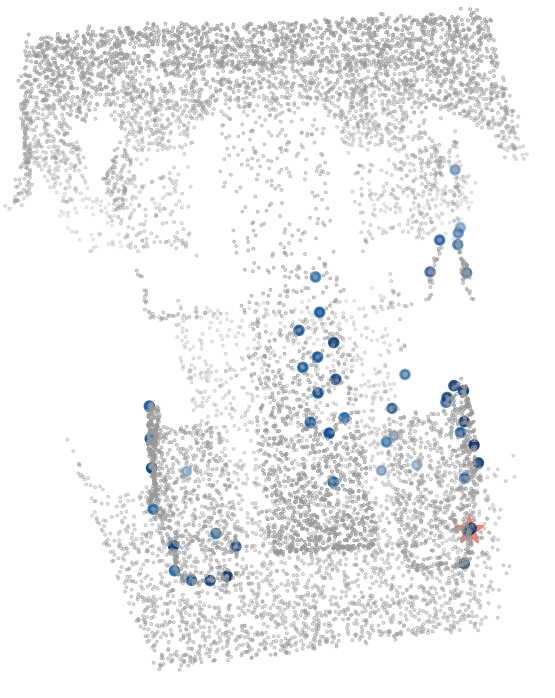} &
    \includegraphics[width=0.14\linewidth]{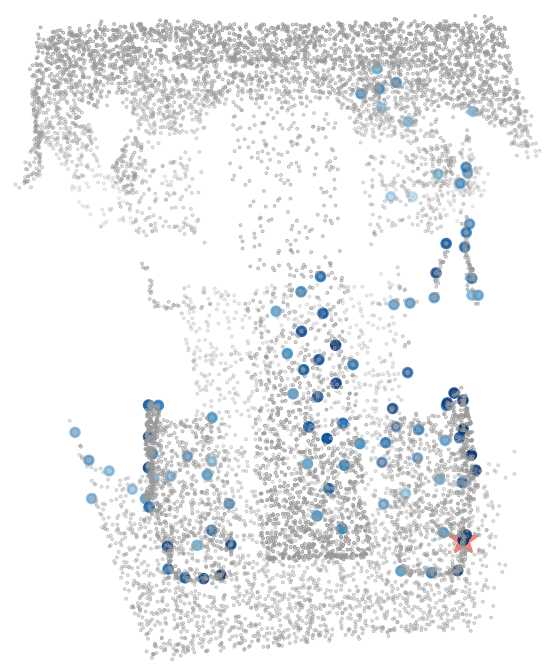} & \includegraphics[width=0.14\linewidth]{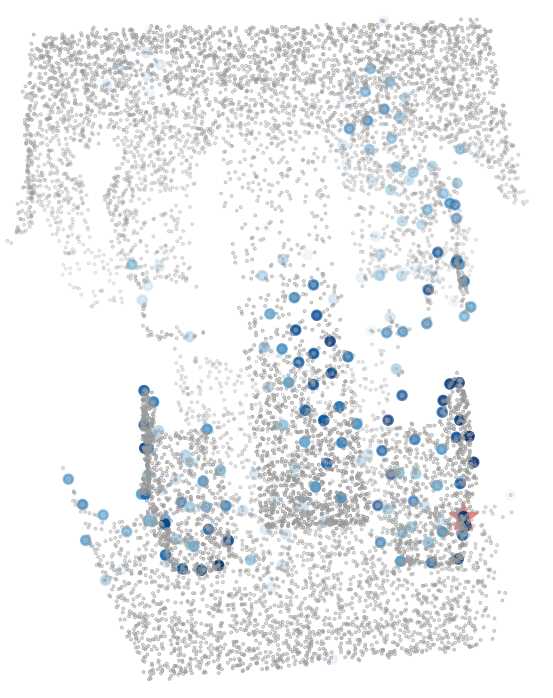} & \includegraphics[width=0.14\linewidth]{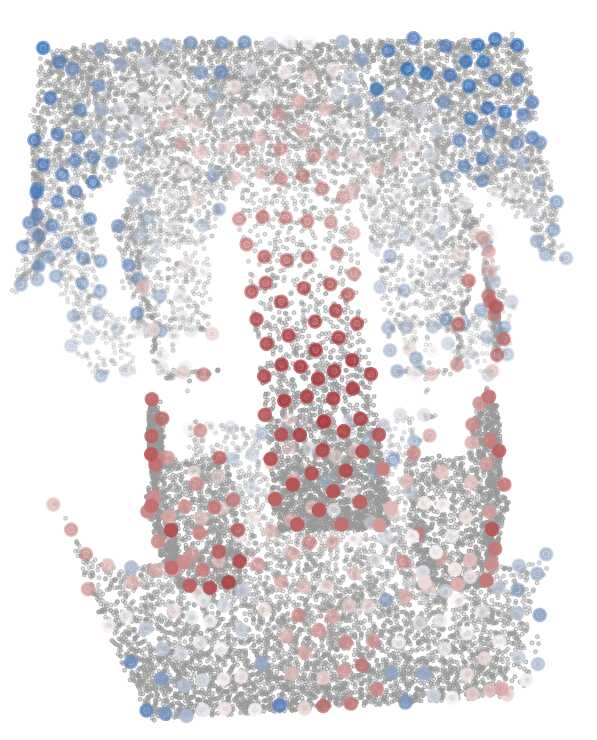} \\
    \includegraphics[width=0.14\linewidth]{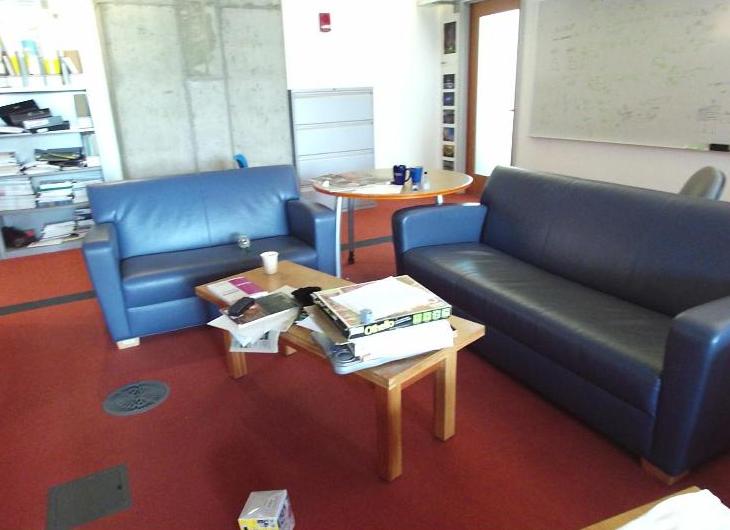} & \includegraphics[width=0.14\linewidth]{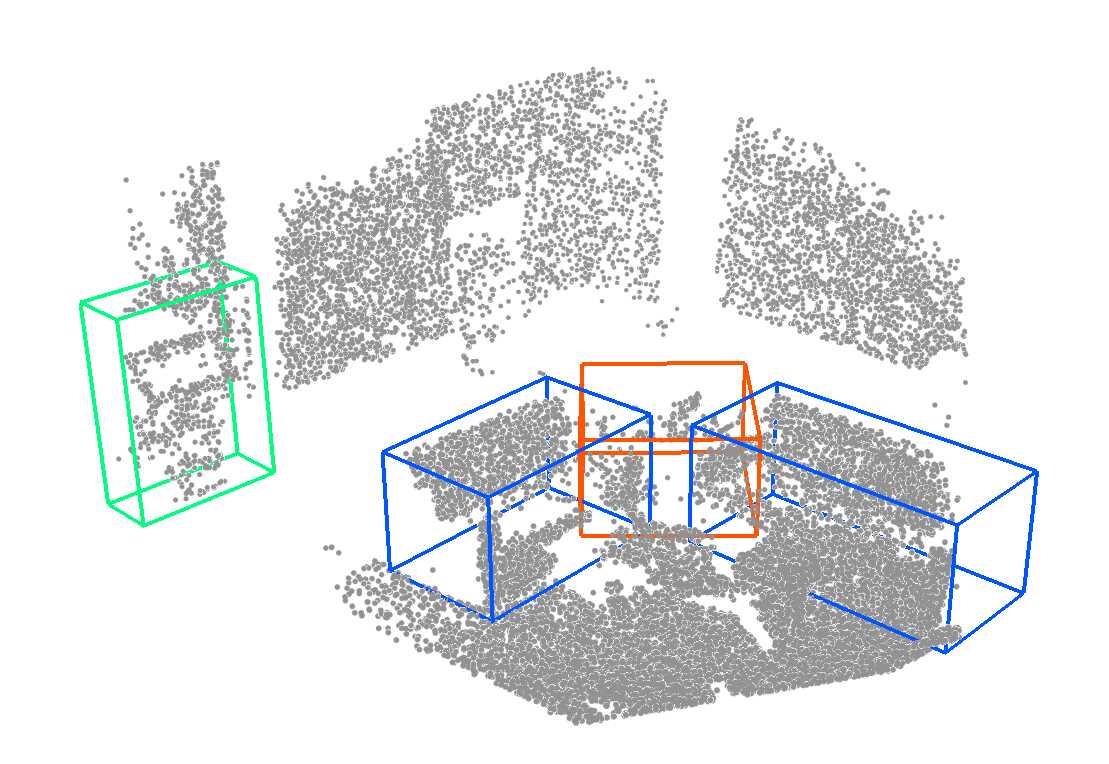} & \includegraphics[width=0.14\linewidth]{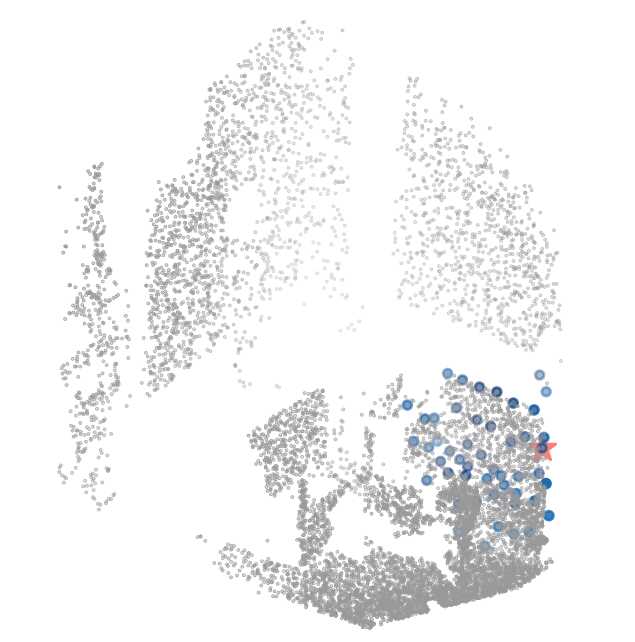} &
    \includegraphics[width=0.14\linewidth]{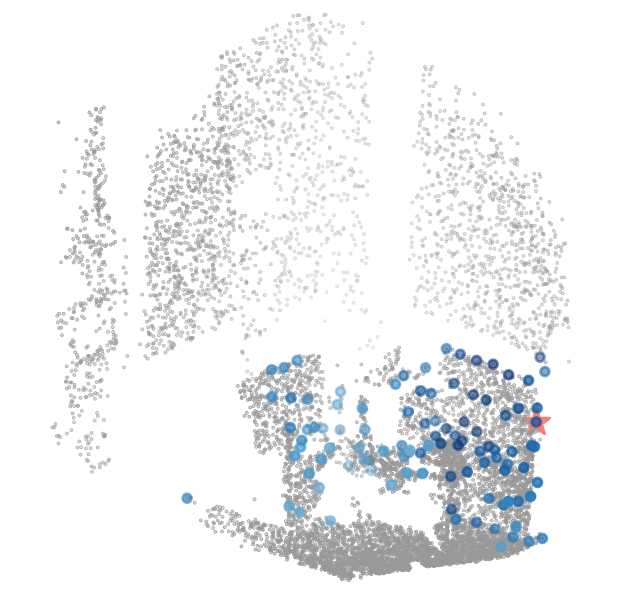} & \includegraphics[width=0.14\linewidth]{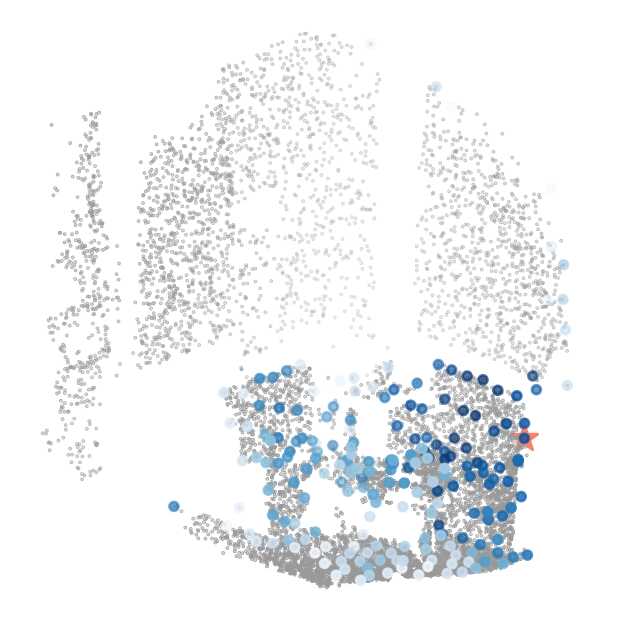} & \includegraphics[width=0.14\linewidth]{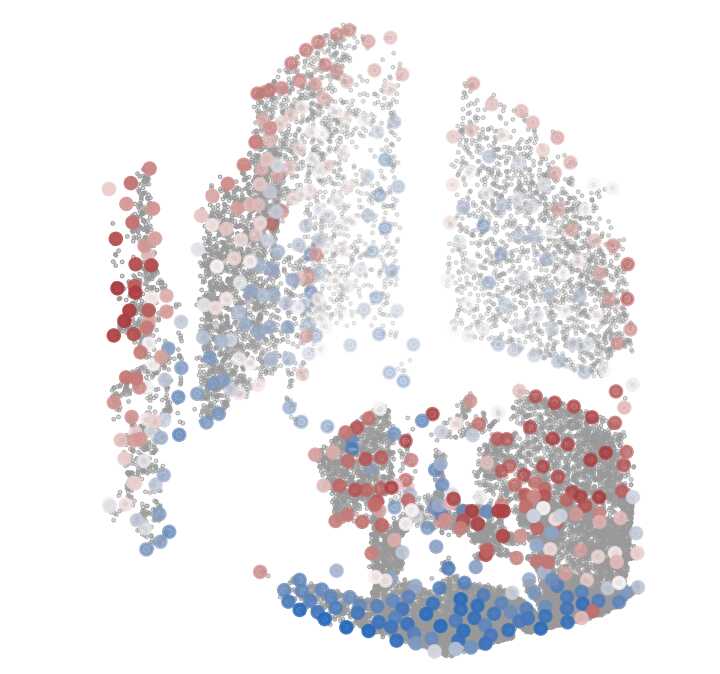} \\
\end{tabular}
\end{center}
\caption{\textbf{More attention maps visualizations on SUN RGB-D.} From left to right: Original scene image, ground truth annotations, top-50, 100, 200 attention maps of points to a query point, and the overall attention map for the entire scene. In top-k attention, the star in orange indicates the query point and darker color indicates larger attention weight, in overall attention red indicates large value. Different object categories are presented with different colors.}
\label{vis_rgbd_att}
\end{figure*}

\begin{figure*}
    \begin{center}
    \begin{tabular}{ccc}
        Scene &
        Predictions &
         Ground truth \\
         \includegraphics[width=0.25\linewidth]{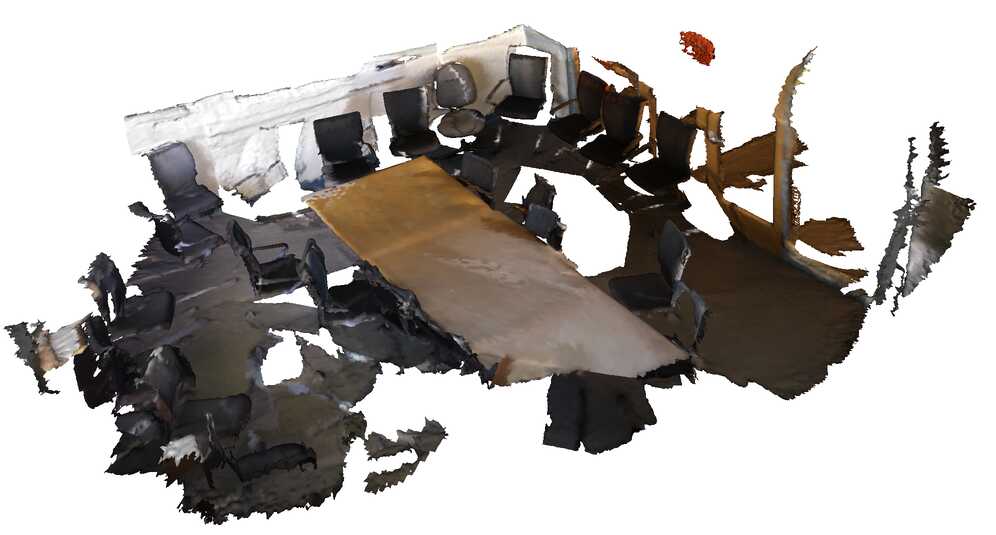} &
        \includegraphics[width=0.25\linewidth]{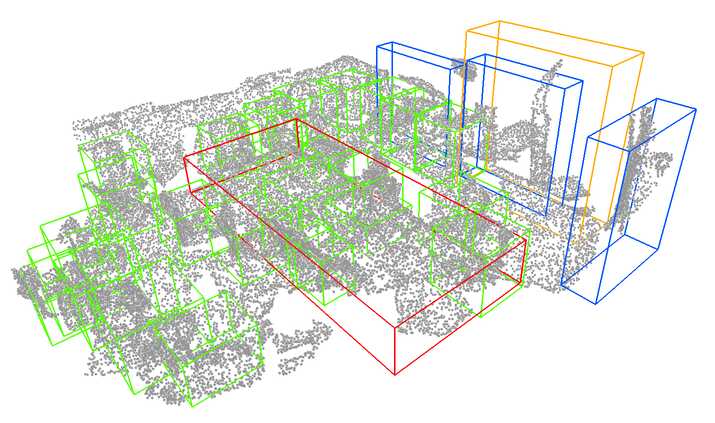} & 
        \includegraphics[width=0.25\linewidth]{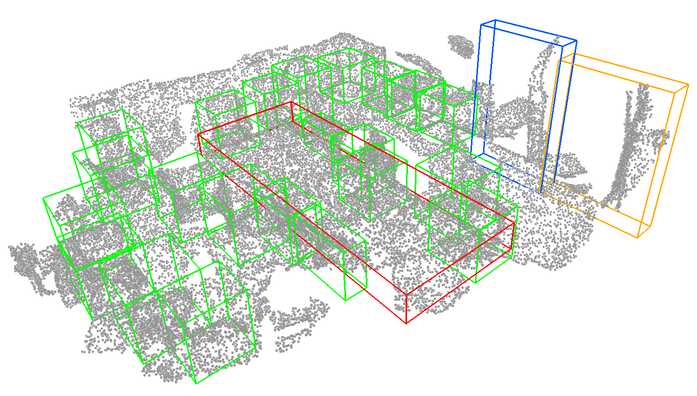}\\
        \includegraphics[width=0.25\linewidth]{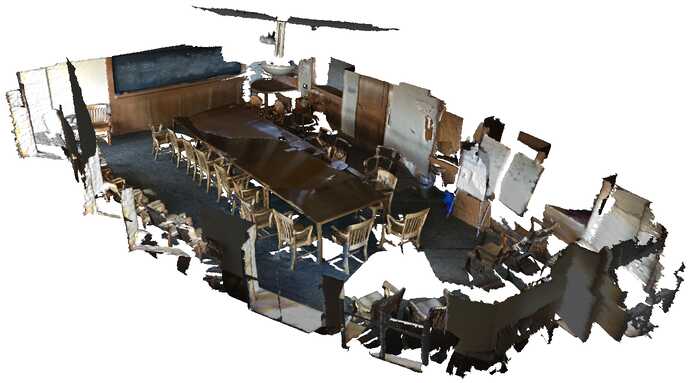} &
        \includegraphics[width=0.25\linewidth]{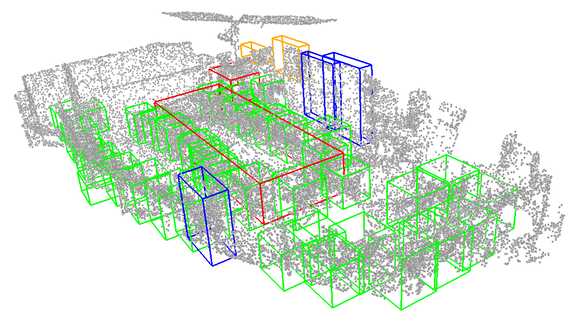} & 
        \includegraphics[width=0.25\linewidth]{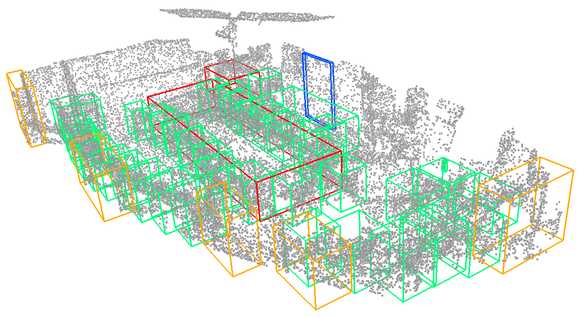}\\
        \includegraphics[width=0.25\linewidth]{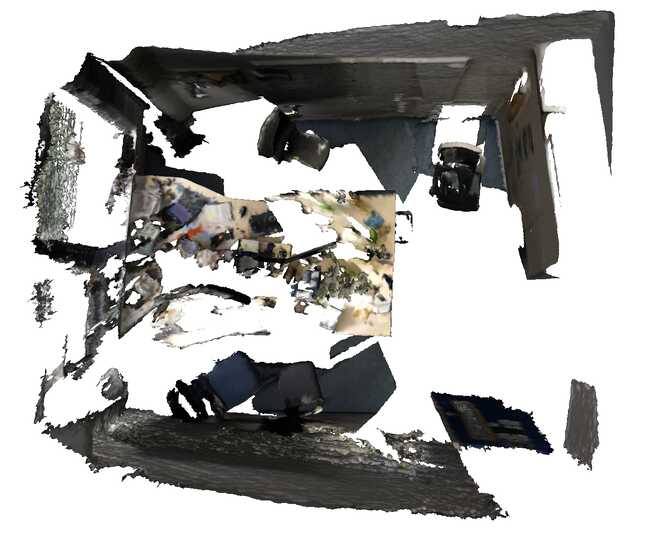} &
        \includegraphics[width=0.25\linewidth]{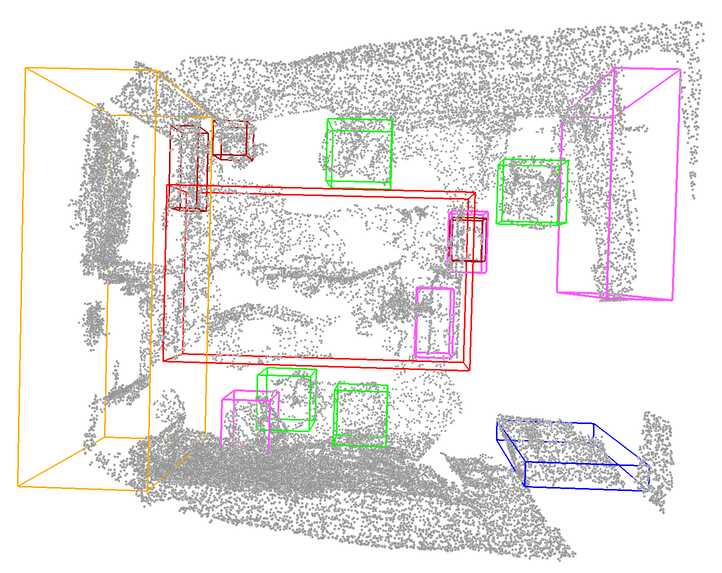} & 
        \includegraphics[width=0.25\linewidth]{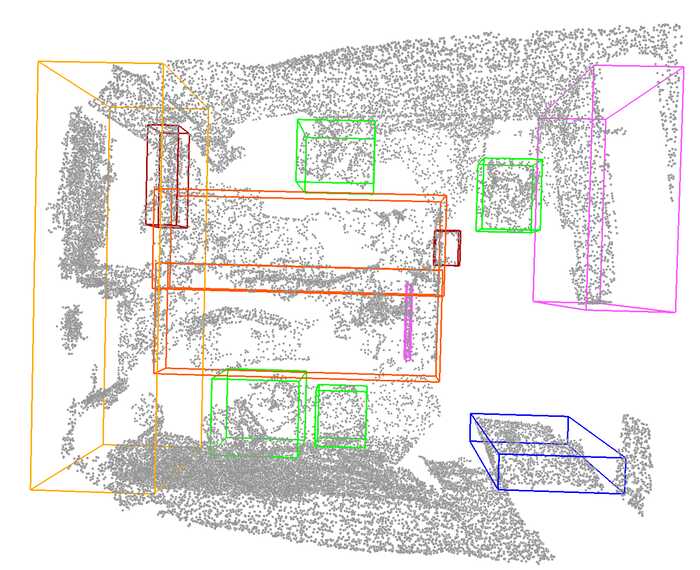}\\
    \end{tabular}
    \end{center}
    \caption{\textbf{Qualitative results of 3D object detection on ScanNetV2.} From left to right: Original scene image, our model's prediction, and annotated ground truth boxes. Different object categories are presented with different colors.}
    \label{vis_scan_det}
\end{figure*}

\begin{figure*}
    \begin{center}
    \begin{tabular}{cc}
         \includegraphics[width=0.45\linewidth]{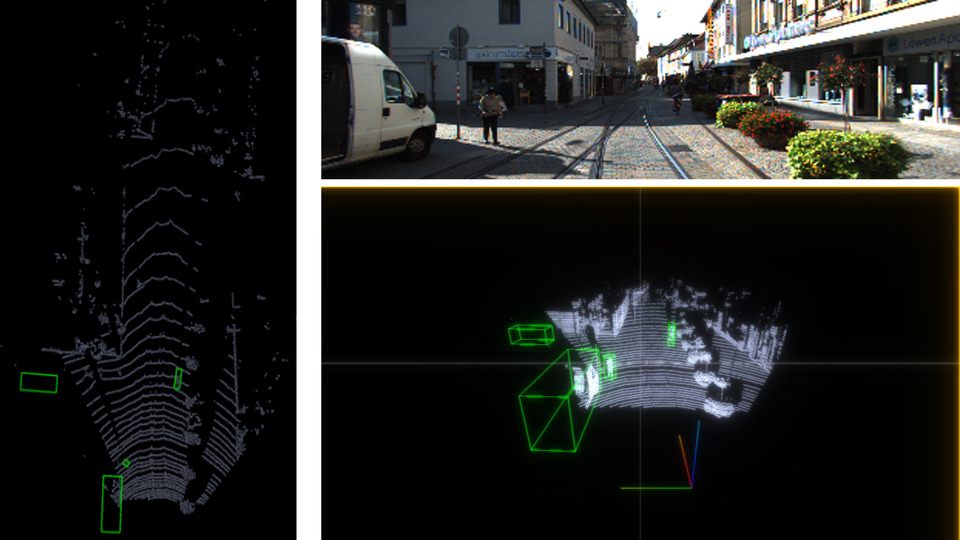} &
         \includegraphics[width=0.45\linewidth]{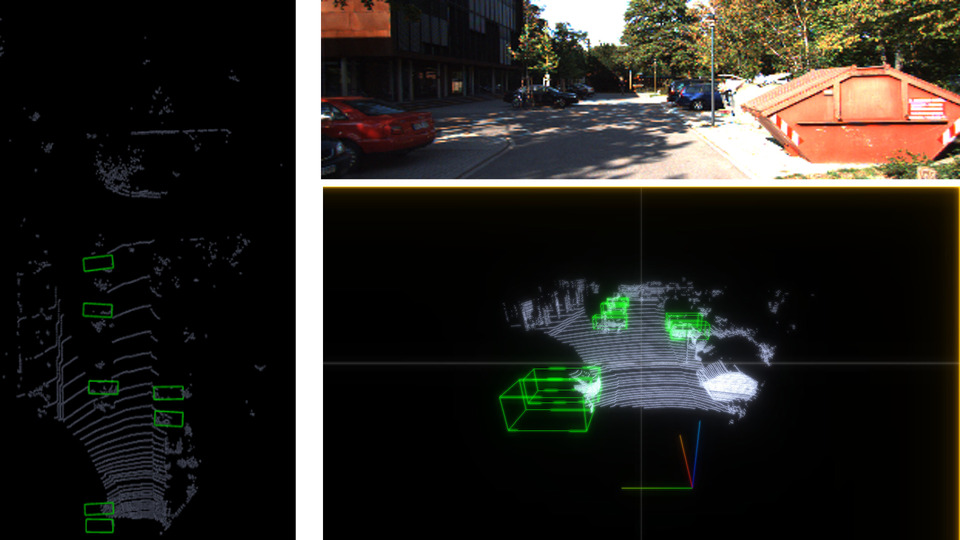} \\
         \includegraphics[width=0.45\linewidth]{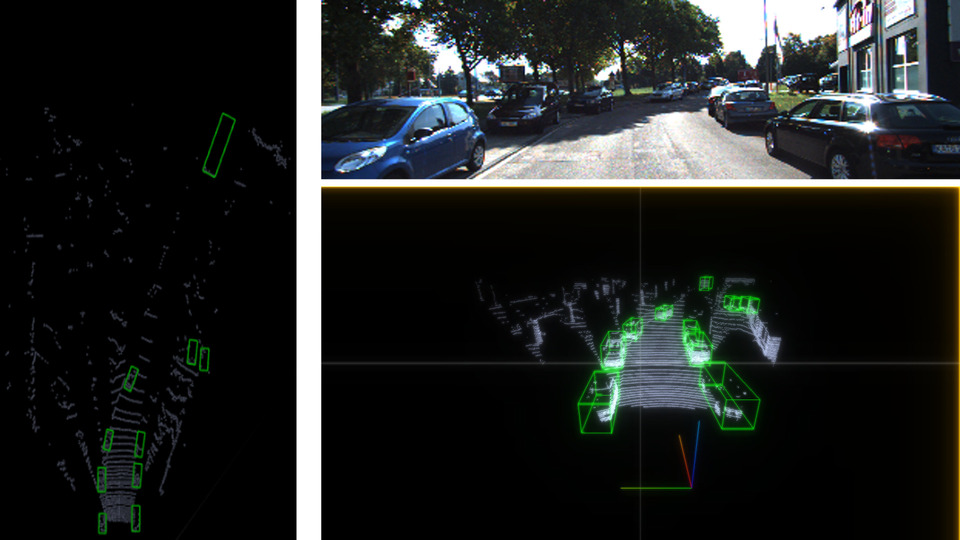} &
         \includegraphics[width=0.45\linewidth]{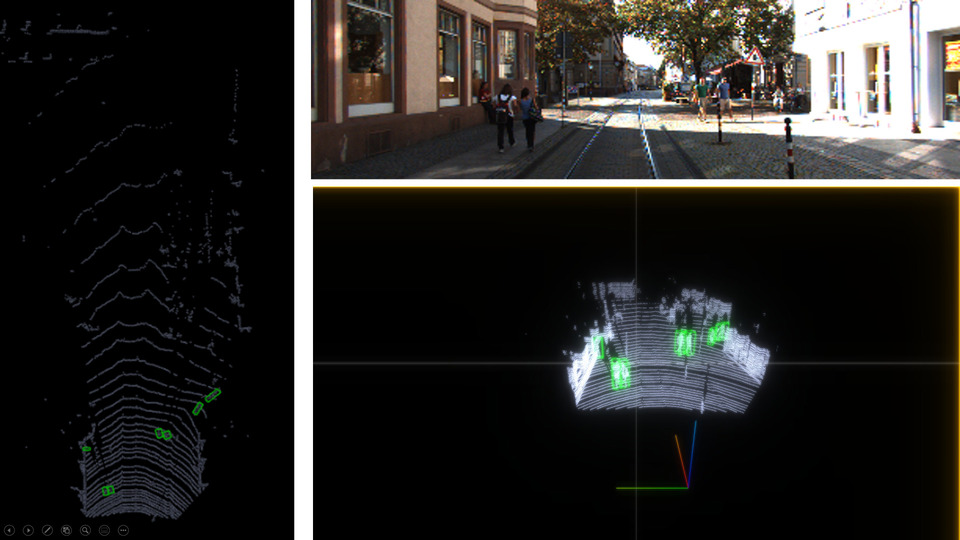} \\
    \end{tabular}
    \end{center}
    \caption{\textbf{Qualitative results of 3D object detection on KITTI val split.} We show detection results in four scenes. In each scene, the left is bird eye view detection results, the upper right is the scene image, and the lower right is the front view detection results. Our detection results are consistent with the ground truth labels (not shown).}
    \label{kitti}
\end{figure*}

We provide additional visualization results in this section. Figure \ref{vis_rgbd_att} shows more visualized attention maps on SUN RGB-D dataset. Figure \ref{vis_scan_det} and Figure \ref{kitti} present qualitative results of detection models with Pointformer on ScanNetV2 and KITTI dataset, respectively.

\end{appendix}


\end{document}